\documentclass[sigconf, screen]{acmart}
\AtBeginDocument{%
  }

\setcopyright{acmlicensed}
\copyrightyear{2025}
\acmYear{2025}
\acmDOI{XXXXXXX.XXXXXXX}
\acmConference[ACM MM '25]{Proceedings of the 2025 ACM International Conference on Multimedia}{October 27--31, 2025}{Dublin, Ireland}
\acmISBN{978-1-4503-XXXX-X/2018/06}

\usepackage{graphicx}
\usepackage{multirow}
\usepackage{array}
\usepackage{subfigure} 
\usepackage{breqn}

\begin{document}

\title{Towards Explainable Fusion and Balanced Learning in Multimodal Sentiment Analysis}


\author[author1]{Miaosen Luo}
\affiliation{%
  \institution{School of Computer Science, South China Normal University}
  \city{Guangzhou}
  \state{Guangdong}
  \country{China}}
\email{luomiaosen@m.scnu.edu.cn}
\author[author2]{Yuncheng Jiang}
\affiliation{%
  \institution{School of Computer Science, South China Normal University}
  \city{Guangzhou}
  \state{Guangdong}
  \country{China}}
\email{ycjiang@scnu.edu.cn}
\author{Sijie Mai}
\authornote{Corresponding author}
\affiliation{%
  \institution{School of Computer Science, South China Normal University}
  \city{Guangzhou}
  \state{Guangdong}
  \country{China}}
\email{sijiemai@m.scnu.edu.cn}

\renewcommand{\shortauthors}{Miaosen Luo et al.}

\begin{abstract}

    Multimodal Sentiment Analysis (MSA) faces two critical challenges: the lack of interpretability in the decision logic of multimodal fusion and modality imbalance caused by disparities in inter-modal information density. To address these issues, we propose KAN-MCP, a novel framework that integrates the interpretability of Kolmogorov-Arnold Networks (KAN) with the robustness of the Multimodal Clean Pareto (MCPareto) framework. First, KAN leverages its univariate function decomposition to achieve transparent analysis of cross-modal interactions. This structural design allows direct inspection of feature transformations without relying on external interpretation tools, thereby ensuring both high expressiveness and interpretability. Second, the proposed MCPareto enhances robustness by addressing modality imbalance and noise interference. Specifically, we introduce the Dimensionality Reduction and Denoising Modal Information Bottleneck (DRD-MIB) method, which jointly denoises and reduces feature dimensionality. This approach provides KAN with discriminative low-dimensional inputs to reduce the modeling complexity of KAN while preserving critical sentiment-related information. Furthermore, MCPareto dynamically balances gradient contributions across modalities using the purified features output by DRD-MIB, ensuring lossless transmission of auxiliary signals and effectively alleviating modality imbalance. This synergy of interpretability and robustness not only achieves superior performance on benchmark datasets such as CMU-MOSI, CMU-MOSEI, and CH-SIMS v2 but also offers an intuitive visualization interface through KAN’s interpretable architecture. Our code is released on https://github.com/LuoMSen/KAN-MCP.
\end{abstract}

\ccsdesc[500]{Information systems~Multimedia information systems}
\ccsdesc[500]{Computing methodologies~Machine learning}
\ccsdesc[100]{Computing methodologies~Natural language processing}
\ccsdesc[100]{Computing methodologies~Supervised learning by regression}

\keywords{Multimodal Sentiment Analysis, Kolmogorov-Arnold Networks, Information Bottleneck, Imbalanced Multimodal Learning}


\maketitle

\section{Introduction}
    Multimodal Sentiment Analysis (MSA) aims to achieve fine-grained modeling and analysis of human sentiments by integrating multi-source information from text, speech, visual and other modalities \cite{zhang2024comprehensive}. Compared to traditional methods relying on single modalities (e.g., text or audio), MSA captures richer sentimental cues through cross-modal complementarity, such as combining sarcasm detection in speech intonation with subtle facial micro-expression variations \cite{lai2023multimodal}. This technology has shown significant value in affective computing, mental health assessment, and intelligent interaction \cite{kalateh2024systematic}.

    However, while multimodal fusion significantly enhances the performance of the model, the increasing complexity of cross-modal interactions obscures the internal feature fusion logic, resulting in opaque decision-making processes that severely hinder model trustworthiness and optimization \cite{adadi2018peeking}. To improve the interpretability of deep learning models, researchers have proposed two categories of solutions from the perspective of explainable AI \cite{retzlaff2024post}. Traditional post-hoc interpretability methods \cite{madsen2022post}, such as LIME \cite{garreau2020explaining} and SHAP \cite{lundberg2017unified}, indirectly infer model logic through local approximations or feature importance analysis. However, their explanations may suffer from approximation errors and do not directly reflect the actual decision-making process of the model. In contrast, ante-hoc interpretable methods, such as linear models and decision trees, offer structural transparency but lack expressive power for modeling complex nonlinear relationships.

\begin{figure}[H]
        \centering
        \vspace{-0.2cm}
        \includegraphics[width=0.48\textwidth]{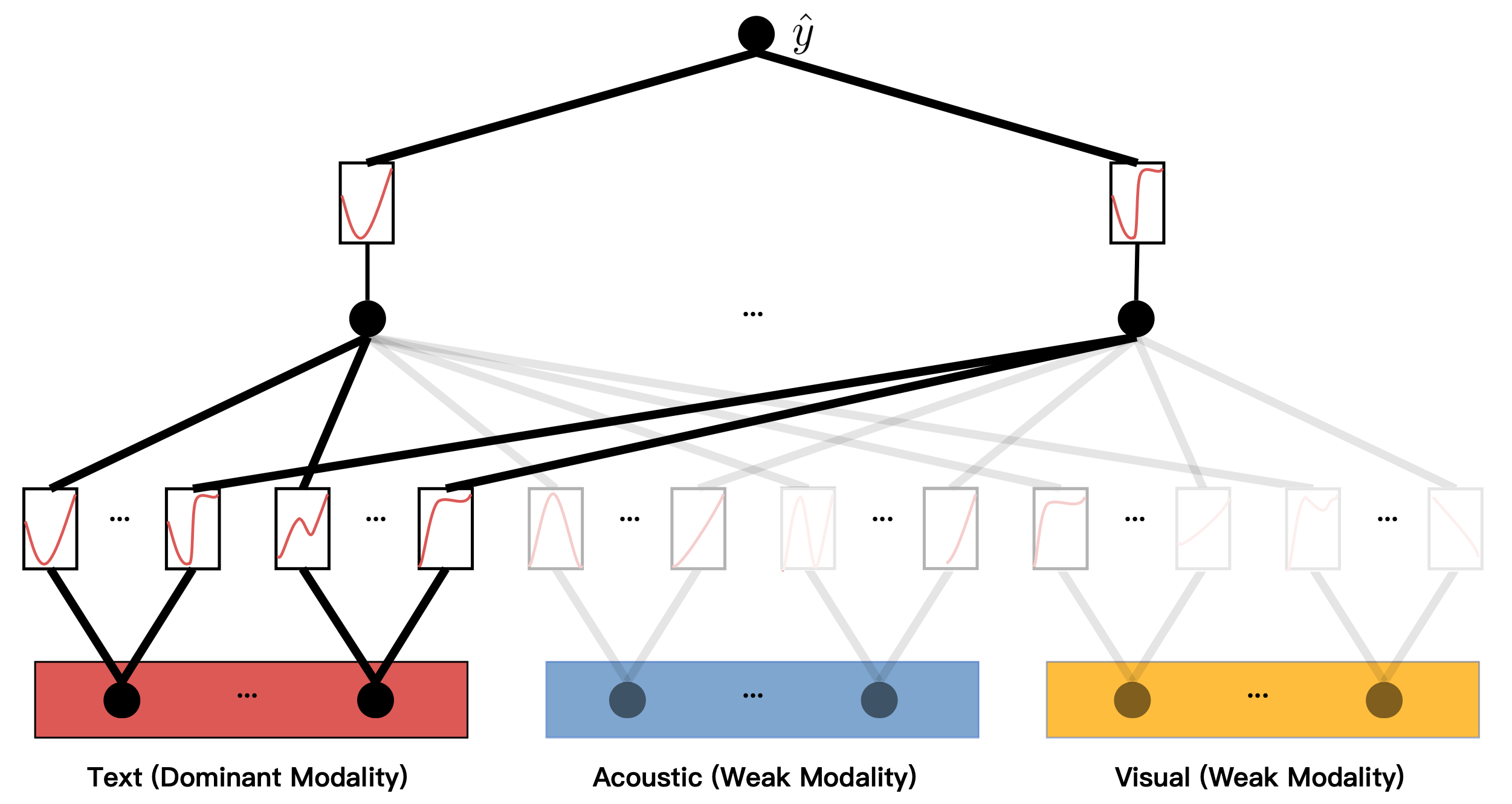} 
        \caption{KAN visualization of the text modality dominance effect for MSA, where the transparency of neuron nodes and connection weights gradually becomes less opaque as the feature contribution levels increase.
        }
        \label{fig:Kolmogorov-Arnold-MSA}
        \vspace{-0.2cm}
    \end{figure}

    To address these limitations, Kolmogorov-Arnold Networks (KAN) \cite{liu2024kan}, based on the Kolmogorov-Arnold representation theorem \cite{schmidt2021kolmogorov}, exhibit unique advantages. The Kolmogorov-Arnold theorem states that any continuous multivariate function can be represented as a finite composition of univariate continuous functions. KAN implements this theory by placing activation functions on the edges (connections) of the network rather than on the nodes. This innovative architecture ensures that each layer comprises univariate functions directly corresponding to input feature transformations, enabling model behavior to be understood by analyzing these univariate functions without relying on external interpretation tools. KAN achieves both high expressiveness (approximating arbitrary continuous functions) and interpretability through structural design, overcoming the performance limitations of traditional ante-hoc methods in complex tasks. Furthermore, KAN’s visualization capability enhances its interpretability. As shown in Figure \ref{fig:Kolmogorov-Arnold-MSA}, by setting the transparency of the activation functions and node connections proportional to their functional magnitudes, users can intuitively identify which data points contribute significantly to the output of the model, thereby focusing on critical features and relationships.

    Despite these advantages, directly applying KAN to MSA remains challenging. When integrating high-dimensional feature vectors from multiple modalities, it is difficult and time-consuming for KAN to analyze the effect of each feature. Moreover, when applying KAN to MSA, we observed that the text modality exhibits significantly higher information density compared to noisy audio and visual modalities \cite{mai2021analyzing,mai2019divide}, causing models to over-rely on text while neglecting contributions from other modalities. To mitigate this, Wei et al. \cite{wei2024mmpareto} analyzed Pareto integration in multimodal scenarios and proposed the multimodal Pareto algorithm (MMPareto). This algorithm aligns gradient directions across all learning objectives to improve generalization and provides harmless unimodal auxiliary signals, alleviating the modality imbalance. However, the original MMPareto algorithm does not account for unimodal noise interference, leading to performance degradation when non-dominant modalities (e.g., audio or visual) contain significant noise.

    Our solution, the Multimodal Clean Pareto (MCPareto) framework, introduces hierarchical optimization. First, in the feature purification layer, we design the Dimensionality Reduction and Denoising Modal Information Bottleneck (DRD-MIB) method based on the Information Bottleneck (IB) principle \cite{tishby2000information, alemi2016deep, wilson2017information, mai2022multimodal}. DRD-MIB maximizes mutual information between features and sentiment labels while filtering irrelevant noise during feature dimensionality reduction. This process not only achieves feature denoising but also adapts to KAN’s univariate function representation capability by reducing feature complexity, ensuring that the model retains minimally sufficient features for effective learning. Subsequently, in the gradient coordination layer, MCPareto leverages the MultiModal Pareto algorithm (MMPareto) \cite{wei2024mmpareto} on the purified unimodal features generated by DRD-MIB. By implementing dual optimization mechanisms—gradient direction alignment and adaptive magnitude modulation—MCPareto reconstructs the multimodal Pareto front. This ensures lossless transmission of unimodal auxiliary signals and significantly enhances generalization robustness through cross-modal gradient field coordination.

    Building on these innovations, we propose KAN-MCP, a framework that synergizes KAN’s interpretability with MCPareto’s robustness. Its core contributions address the aforementioned challenges through three key aspects:
    
    \begin{itemize}
        \item Interpretable Fusion Mechanism: We introduce KAN to the MSA field for the first time, enabling transparent cross-modal interaction analysis through univariate function decomposition and visual pathways.
        \item Noise Robustness Enhancement: The proposed DRD-MIB method jointly denoises and reduces feature dimensionality, providing KAN with discriminative low-dimensional inputs and assisting the alleviation of modality imbalance issue.
        \item Modality-Balanced Learning: The MCPareto strategy dynamically balances gradient contributions using purified features from DRD-MIB, effectively alleviating the modality imbalance based on more discriminative unimodal features.
    \end{itemize}
    
    Experiments demonstrate that our method significantly outperforms existing models on the CMU-MOSI, CMU-MOSEI, and CH-SIMS v2 benchmarks. Additionally, KAN’s visualization interface intuitively reveals decision logic, offering theoretical tools and practical paradigms for trustworthy deployment of multimodal models.

\section{Related Work}
    \subsection{MSA Fusion Methods}
        To comprehensively capture the complexity of human emotions, multimodal information fusion plays a pivotal role in Multimodal Sentiment Analysis (MSA). By integrating features or decisions from diverse modalities such as audio, visual, and textual data, multimodal fusion methods significantly enhance the accuracy of emotion polarity inference. In recent years, researchers have proposed tensor-based fusion \cite{hou2025tf, miao2024low}, attention-based fusion \cite{huang2025atcaf, liu2024sentiment, liu2024transformer}, translation-based fusion \cite{zeng2024disentanglement, liu2024modality, lu2024coordinated}, and context-aware fusion \cite{wang2025cime, lu2024hypergraph} to improve the efficiency of data fusion.

        Despite the notable improvements in emotion recognition accuracy and robustness, these fusion methods introduce new challenges due to their inherent complexity. Intricate fusion mechanisms render the model’s decision-making process difficult to interpret intuitively, reducing transparency in real-world applications and complicating further optimization. Thus, balancing model complexity with interpretability while pursuing higher performance has emerged as a critical challenge in MSA.
        
    \subsection{Interpretability Methods in MSA}

        Traditional explainable Artificial Intelligence (xAI) fall into two categories \cite{retzlaff2024post}: ante-hoc interpretability, which employs structured designs (e.g., feature importance evaluation \cite{wu2022interpretable}, rule-based systems \cite{aghaeipoor2023fuzzy}, attention mechanisms \cite{du2023saits}) to build transparent and interpretable models; and post-hoc interpretability, which reverse-engineers trained models via gradient analysis \cite{zeng2023abs}, perturbation testing, or visualization techniques \cite{wang2020cnn}.


        In the MSA domain, Li et al. \cite{li2021quantum} proposed a quantum-inspired multimodal fusion (QMF) method that models intra-modal feature correlations through quantum superposition, fuses cross-modal decisions via quantum entanglement, and disentangles single-/bi-modal contributions using reduced density matrices. While this approach provides explicit explanations for modal interactions, its quantum state abstraction lacks intuitive mapping to practical features. Wu et al. \cite{wu2022interpretable} introduced Interpretable Multimodal Capsule Fusion (IMCF), which explicitly quantifies modal importance via routing coefficients but fails to reveal cross-layer feature transformation logic. Tsai et al. \cite{tsai2020multimodal} developed a multimodal routing method that achieves local explanations through dynamic weight allocation, yet its interpretability remains confined to numerical weight levels. Khalane et al. \cite{khalane2025evaluating} applied Gradient SHAP \cite{lundberg2017unified} to quantify feature contributions, but local gradients cannot capture the global impact of complex multimodal interactions. Nie et al. \cite{nie2024interpretable} utilized integrated gradients to visualize input feature contributions, yet their approach inadequately explains internal decision paths and cross-modal synergies.

        This work is the first to introduce KAN into MSA. By explicitly decomposing multimodal interactions via B-spline-parameterized edge functions, KAN transforms traditional "black-box" feature fusion into mathematically analyzable compositions of univariate functions. This functional decomposition mechanism not only visualizes multimodal decision pathways but also quantifies the dynamic contributions of different modalities during feature space transformations, offering a novel theoretical framework and technical pathway for resolving interpretability challenges in multimodal fusion. Unlike traditional methods limited to local explanations or abstract representations, KAN ensures mathematical verifiability across the entire pipeline from input features to final decisions, establishing a fine-grained interpretability system while maintaining model performance.

    \subsection{Feature Reduction Methods}
        In deep learning, feature reduction methods can be categorized into supervised (e.g., CNN/Transformer), unsupervised (e.g., autoencoders/DBN), and semi-supervised (e.g., pseudo-label pretraining) \cite{jia2022feature}. Traditional supervised methods rely on implicit backpropagation for feature selection, lacking theoretical frameworks for noise separation. Unsupervised methods tend to retain task-irrelevant information due to the absence of label guidance. Semi-supervised methods suffer from cumulative pseudo-label noise. Although the Information Bottleneck (IB) theory provides an information-theoretic foundation for feature compression, its application in deep learning often faces limitations such as insufficient multimodal noise suppression and high computational complexity. The proposed DRD-MIB method addresses these issues through targeted mutual information optimization, explicitly filtering single-modal noise during dimensionality reduction while introducing inter-modal mutual information constraints to prevent cross-modal conflicts.

    \subsection{Imbalanced Multimodal Learning}
    
        In our study, the imbalance problem in multimodal learning poses a critical challenge, as it leads models to overly rely on dominant modalities while neglecting others. To address this, Peng et al. \cite{peng2022balanced} proposed the OGM-GE algorithm, which dynamically adjusts gradient updates across modalities. Wu et al. \cite{wu2022characterizing} introduced a conditional learning velocity method to balance learning processes by capturing relative learning speeds among modalities. Guo et al. \cite{guo2024classifier} developed a Classifier-Guided Gradient Modulation (CGGM) approach that employs modality-specific classifiers to evaluate contributions and dynamically modulate gradient magnitudes.

        However, these methods exhibit limitations in specific scenarios. For instance, CGGM \cite{guo2024classifier} requires additional classifiers for gradient modulation and mandates fully connected layers for both single-modal and multi-modal classifiers, making it incompatible with our KAN-based architecture. The conditional learning velocity method by Wu et al. \cite{wu2022characterizing}, designed for intermediate fusion strategies, struggles to scale to scenarios with more than two modalities. OGM-GE \cite{peng2022balanced} focuses solely on gradient magnitude adjustment while ignoring directional alignment.

        To overcome these issues, Wei et al. \cite{wei2024mmpareto} proposed the MMPareto algorithm, which leverages multi-objective Pareto optimization to balance both gradient direction and magnitude during updates. MMPareto eliminates the need for auxiliary classifiers or specific architectures and supports arbitrary numbers of modalities, aligning well with our requirements. However, it fails to address the impact of intra-modal noise on Pareto frontiers. Our proposed MCPareto adopts a hierarchical optimization framework comprising two layers: (1) a feature purification layer that employs DRD-MIB to suppress single-modal noise and compress dimensions, filtering out emotion-unrelated interference; and (2) a gradient coordination layer that synergizes denoised features with MMPareto to significantly enhance the utilization of minor modalities (e.g., audio and visual), thereby improving overall model performance.

\begin{figure*}[ht]
    \centering
    \includegraphics[width=0.9\textwidth]{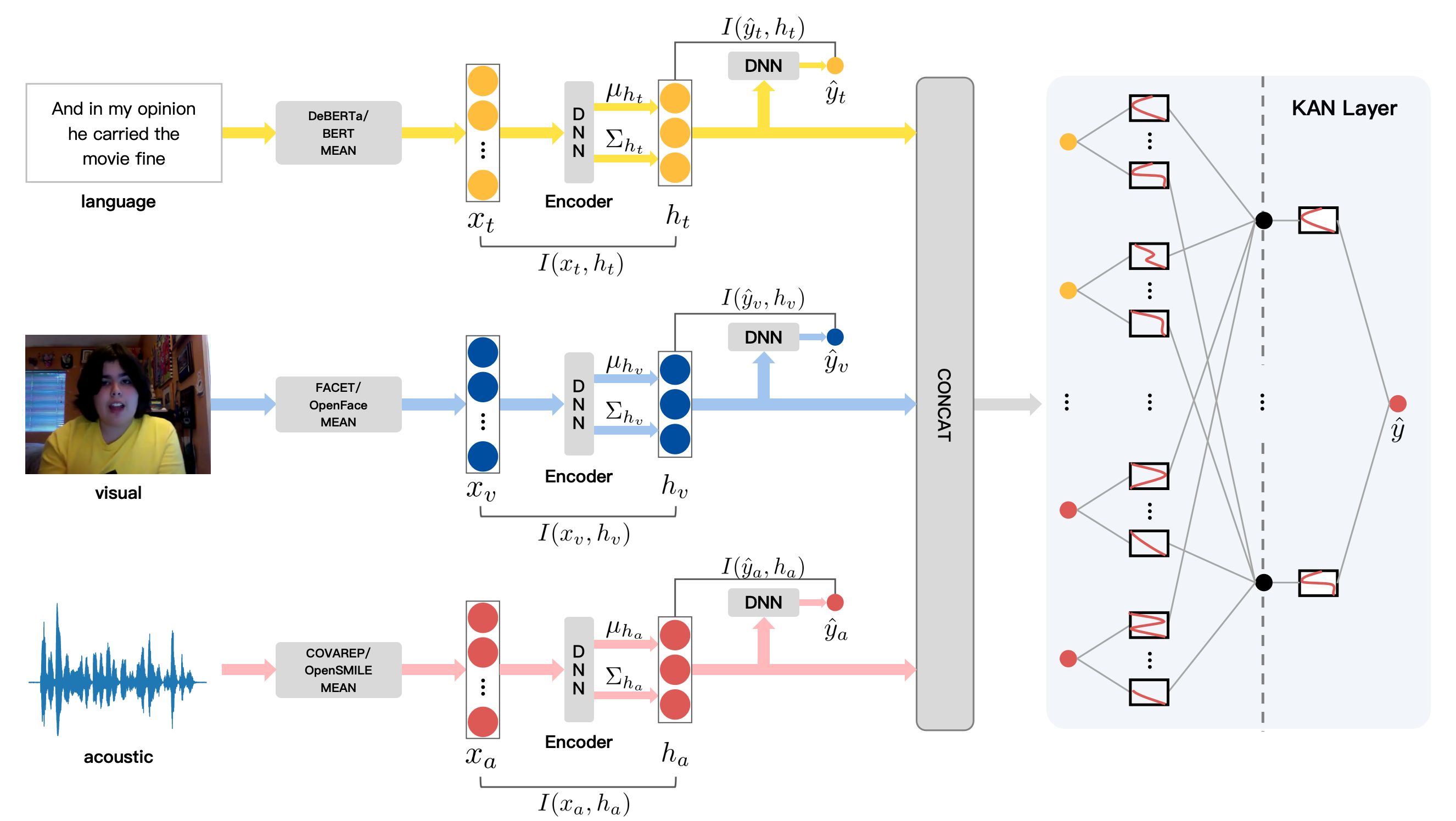} 
    \caption{The overall architecture of the proposed KAN-MCP.}
    \label{fig:KAN-MCP}
\end{figure*}

\section{Model Architecture}
    In this section, we present the detailed architecture of KAN-MCP, a novel framework that integrates the interpretability of KAN with the robustness of the MCPareto framework. As shown in Figure \ref{fig:KAN-MCP}, the proposed model addresses the challenges of interpretability and multimodal learning imbalance in MSA through a two-stage design: a feature encoding stage and a fusion-prediction stage. Below, we elaborate on each component and their synergistic integration.
    
\subsection{Encoding Stage}
    The encoding stage is responsible for extracting and purifying features from linguistic, acoustic, and visual modalities. This stage addresses the issue of modality noise interference through the Dimensionality Reduction and DRD-MIB method, which is based on the IB) principle. DRD-MIB ensures that only the most relevant information is retained while filtering out noise, thereby providing clean and compact representations for downstream tasks.
    
    \subsubsection{Unimodal Feature Extraction} We first perform feature extraction on each modality separately to obtain latent representations. For the textual modality, we leverage pre-trained language models (e.g., DeBERTa \cite{he2020deberta} or BERT \cite{devlin2019bert}) to tokenize and encode textual input $I_t$ into a feature sequence $x_t\in \mathbb R^{d_t\times T_t}$. For the acoustic and visual modalities, we use preprocessed feature sequences $x_a\in \mathbb R^{d_a\times T_a}$ and $x_v\in \mathbb R^{d_v\times T_v}$, respectively. Temporal averaging is then applied to derive modality-specific representations.
    
    \subsubsection{Unimodal Feature Reduction}
    To filter out noise and redundancy while achieving feature dimensionality reduction, we apply DRD-MIB to encode $x_m$ into a minimally sufficient predictive representation $h_m$. DRD-MIB enables unimodal representations to filter out noisy information prior to fusion and aligns the encoded unimodal distributions by individually applying the Information Bottleneck (IB) principle to each modality. The objective of DRD-MIB is formulated as:
    
    \begin{equation}
        \mathcal{L}_{DRD-MIB} = \mathcal{I}(y;h) + \sum_{m\in\{t,a,v\}}[\mathcal{I}(y;h^m)-\gamma \mathcal{I}(x^m; h^m)]
    \end{equation}
    where $y$ is the label, $h$ is the fused multimodal representaion, $x^m$ denotes the unimodal representation output by the unimodal learning network, and $h^m$ is the encoded unimodal representation.

    DRD-MIB can be approximated by the following objective function. For brevity, the detailed derivation is relegated to the Appendix, and we present only the final result here:

    \begin{equation}
        \scalebox{0.8}{$
            \begin{aligned}
                \mathcal{J}_{DRD-MIB} &\approx \frac{1}{b} \sum_{i=1}^{b} \biggl[ \log \mathcal{Q}(y_i \mid \boldsymbol{h}_i) + \\&\sum_{m\in\{t,a,v\}} \left[ \mathbb{E}_{\boldsymbol{\epsilon} \sim \mathbb{P}(\boldsymbol{\epsilon})} \left[ \log \mathcal{Q}(y_i \mid \boldsymbol{h}_i^m) \right] - \gamma \cdot KL \left( \mathbb{P}(\boldsymbol{h}_i^m \mid \boldsymbol{x}_i^m) \, \bigg| \, \bigg| \, q(\boldsymbol{h}_i^m) \right) \right] \biggr] \\&= \frac{1}{b} \sum_{i=1}^{b} \biggl[ \log \mathcal{Q}(y_i \mid \boldsymbol{h}_i) + \sum_{m\in\{t,a,v\}} \left[ \mathbb{E}_{\boldsymbol{\epsilon} \sim \mathbb{P}(\boldsymbol{\epsilon})} \left[ \log \mathcal{Q}(y_i \mid \boldsymbol{h}_i^m) \right] \right. \\&\left. - \beta \cdot KL \left( \mathcal{N}(\boldsymbol{\mu}_{h_i^m}, \boldsymbol{\Sigma}_{h_i^m}) \, \bigg| \, \bigg| \, \mathcal{N}(0, \mathcal{I}) \right) \right] \biggr]
            \end{aligned}
        $}
    \end{equation}
    where,
    \begin{equation}
    \begin{aligned}
     \mathbb{P}(h^m \mid x^m) &= \mathcal{N}(\mu^m(x^m; \theta_\mu^m), \Sigma(x^m; \theta_\Sigma^m)) \\
        &= \mathcal{N}(\mu_{h^m}, \Sigma_{h^m})
    \end{aligned}
    \end{equation}
    \begin{equation}
    h^m = \mu_{h^m} + \Sigma_{h^m} \times \epsilon
    \end{equation}
    where $b$ is the sampling size (i.e., batch size), and $i$ is the subscript that indicates each sample, $\mu^m$ and $\Sigma^m$ are the deep neural networks that learn the mean and the variance of $\mathbb{P}(h^m|x^m)$ respectively. From the perspective of filtering out noisy information before fusion, DRD-MIB is effective as it forces the encoded unimodal representation $h^m$ to only contain the relevant information to the target. Moreover, by using the reparameterization trick, the distributions of different encoded unimodal representations are closer, which narrows down the modality gap.

\subsection{Fusion-Prediction Stage}
    The fusion-prediction stage leverages KAN to achieve interpretable and efficient multimodal fusion. Unlike traditional neural networks, KAN places activation functions on the edges (connections) rather than nodes, enabling it to approximate complex multivariate functions through compositions of univariate functions. This design enhances interpretability by allowing direct analysis of feature transformations without external interpretation tools. Specifically, we processes the compressed unimodal features $h_t,h_a,h_v$ and concatenates them into a fused vector $F$, enabling efficient and interpretable representation for downstream tasks.
    Subsequently, KAN processed this fused representation to generate the final prediction:
    \begin{equation}
        \hat y = \text{KAN}(F)
    \end{equation}
    
    To better understand how KAN achieves this, we introduce its unique architecture, which is inspired by the Kolmogorov-Arnold Representation Theorem. Unlike traditional multi-layer perceptrons (MLPs), KAN employs learnable activation functions on edges (weights rather) than fixed activations on nodes (neurons). This design not only enhances interpretability but also achieves higher accuracy with fewer parameters. Formally, KAN is defined as:
    \begin{equation}
        \text{KAN}(x) = (\Phi_{L-1} \circ \Phi_{L-2} \circ \cdots \circ \Phi_1 \circ \Phi_0)(x)
    \end{equation}
    where $\Phi_i$ denotes the $i$-th layer. Each layer with $n_{in}$ inputs and $n_{out}$ outputs is represented as a matrix of univariate functions:
    \begin{equation}
        \Phi = \{\phi_{q,p}\},\quad p=1,2,\cdots,n_{in},\quad q=1,2,\cdots,n_{out}
    \end{equation}
    Here, $\phi_{q,p}$ is a univariate function with trainable parameters. The transformation from layer $l$ to $l+1$ is expressed as:
    \begin{equation}
        \mathbf{x}_{l+1} = \begin{pmatrix}\phi_{l,1,1}(\cdot) & \phi_{l,1,2}(\cdot) & \cdots & \phi_{l,1,n_l}(\cdot) \\\phi_{l,2,1}(\cdot) & \phi_{l,2,2}(\cdot) & \cdots & \phi_{l,2,n_l}(\cdot) \\\vdots & \vdots & \ddots & \vdots \\\phi_{l,n_{l+1},1}(\cdot) & \phi_{l,n_{l+1},2}(\cdot) & \cdots & \phi_{l,n_{l+1},n_l}(\cdot)\end{pmatrix}\mathbf{x}_l,
    \end{equation}
    This structure allows KAN to approximate complex functions via univariate function compositions, enhancing multimodal fusion and prediction performance. By employing edge-based activation functions, KAN achieves interpretability and efficiency, overcoming traditional neural networks' limitations.

\begin{figure}[t]
        \centering
        \includegraphics[width=0.48\textwidth]{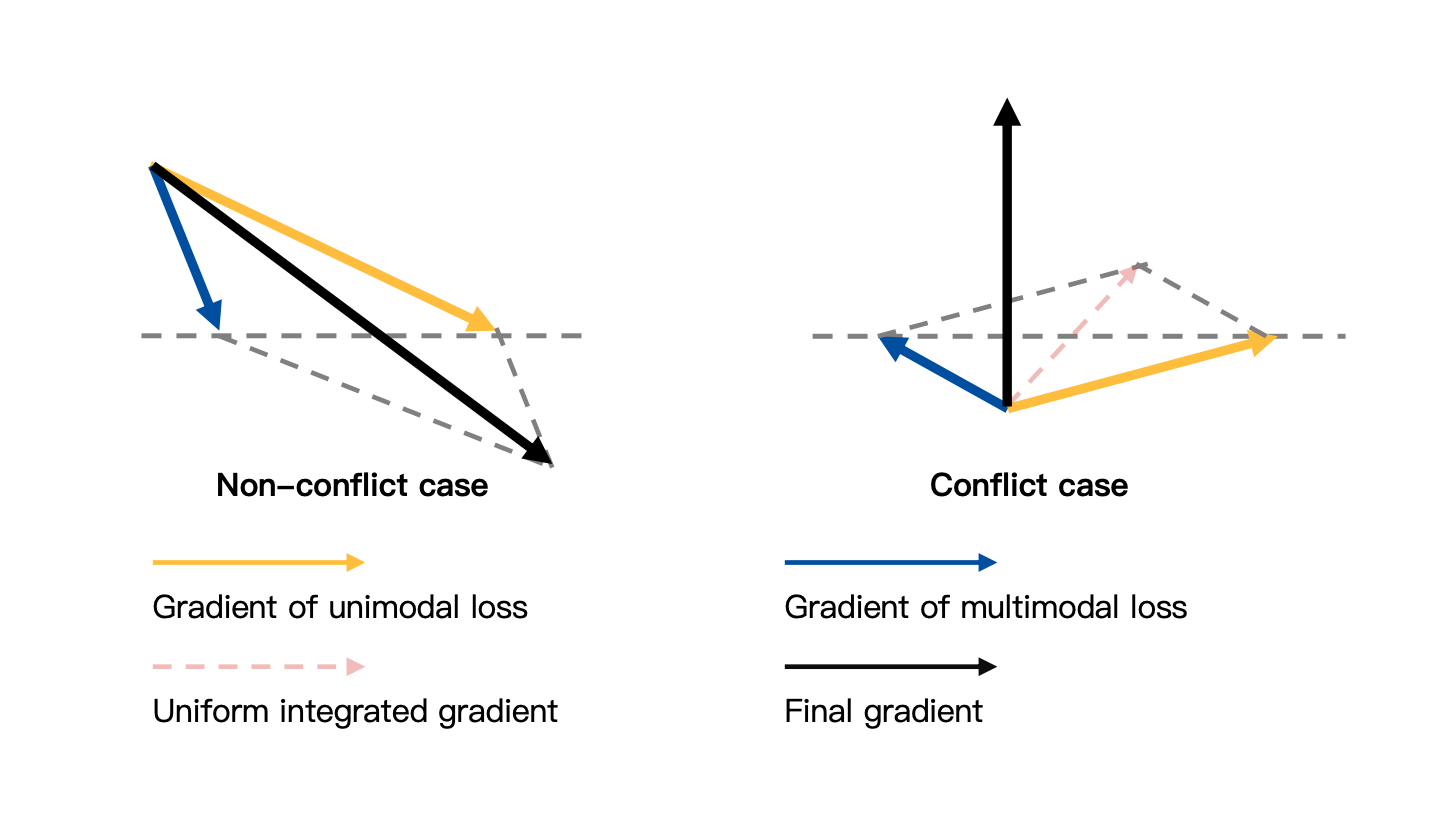} 
        \vspace{-0.6cm}
        \caption{Gradient integration strategy of MMPareto.}
        \label{fig:MMPareto}
        \vspace{-0.2cm}
    \end{figure}
    
\subsection{Multimodal Learning Imbalance}
        

    This study proposes the MCPareto framework to address modality imbalance and noise interference in multimodal learning. To overcome the limitations of MMPareto algorithm that neglect noise in non-dominant modalities leading to performance degradation, we adopt a two-stage optimization approach: First, the DRD-MIB algorithm reduces single-modality noise while enhancing reliable gradients from non-dominant modalities. Subsequently, an MMPareto optimization algorithm coordinates gradient balance between single-modality and multimodal objectives. This dual-stage collaborative mechanism effectively enhances model robustness and comprehensive performance, overcoming the constraints of existing methods. For our task of simultaneously optimizing single-modality and multimodal gradients, the Pareto algorithm can be formulated as follows:
    
    \begin{equation}
        \begin{aligned}
        \min_{\alpha^m, \alpha^u \in \mathbb{R}} & \quad \|\alpha^m \mathbf{g}_S^m + \alpha^u \mathbf{g}_S^u\|^2 \\
        \text{s.t.} & \quad \alpha^m, \alpha^u \geq 0, \\
        & \quad \alpha^m + \alpha^u = 1,
        \end{aligned}
    \end{equation}

    where $g^m_S$ and $g^u_s$ denote the gradients of the multimodal loss and the unimodal loss over a minibatch, respectively. By computing the cosine similarity $\cos\beta$ between $g^m_S$ and $g^u_s$, potential conflicts can be detected. The MMPareto algorithm is illustrated in Figure \ref{fig:MMPareto}.

    Under non-conflicting conditions, the directions of gradient vectors are aligned, enabling their contributions to jointly benefit all learning objectives. This implies that during optimization, these gradients can simultaneously improve multiple objectives without mutual interference. In such cases, the weights $\alpha^m$ and $\alpha^u$ are both set to 0.5 (instead of using the Pareto analytical solution) to amplify the SGD noise term, thereby enhancing generalization capability. The final gradient is formulated as:

    \begin{equation}
        \text{h}_S^{\text{MMPareto}} \sim \mathcal{N}\left( \mathbf{g}_N^m + \mathbf{g}_N^u, \frac{C^m + C^u}{|S|} \right)
    \end{equation}

    Under conflicting conditions, the optimal weights $\alpha^m$ and $\alpha^u$ are derived by solving the Pareto optimization problem to identify a conflict-free direction. The parameter $\gamma$ is then applied to amplify the SGD noise term. The final gradient is:

    \begin{equation}
        \text{h}_S^{\text{MMPareto}} \sim \mathcal{N}\left( \lambda(\mathbf{g}_N^m + \mathbf{g}_N^u), \lambda^2\frac{(2\alpha^m)^2C^m + (2\alpha^u)^2C^u}{|S|} \right)
    \end{equation}
    where,
    \begin{equation}
        \lambda = \frac{\|\mathbf g_S^m+\mathbf g^u_S\|}{\|2\alpha^m\mathbf g_S^m+2\alpha^u\mathbf g^u_S\|}>1
    \end{equation}

    By integrating the MCPareto framework, our model architecture effectively mitigates imbalance in multimodal learning, ensuring that unimodal auxiliary tasks positively contribute to the primary multimodal learning objective.

\begin{table*}[t]
\caption{The comparison with baselines on the CMU-MOSI dataset and CMU-MOSEI dataset. Apart from ULMD, MFON, ConFEDE and UniMSE, the results of baselines are derived from our experiments. The best results are highlighted in bold, and the runner-up results are indicated with underlines.}
\vspace{-0.2cm}
\label{tab:CMUResult}
\begin{tabular}{l|c|c|c|c|c|c|c|c|c|c}
\hline
\multirow{2}{*}{Model} & \multicolumn{5}{c|}{CMU-MOSI} & \multicolumn{5}{c}{CMU-MOSEI} \\ \cline{2-11}
 & Acc7↑ & Acc2↑ & F1↑ & MAE↓ & Corr↑ & Acc7↑ & Acc2↑ & F1↑ & MAE↓ & Corr↑ \\ \hline
Graph-MFN\cite{zadeh2018multimodal} & 34.4 & 80.2 & 80.1 & 0.939 & 0.656 & 51.9 & 84.0 & 83.8 & 0.569 & 0.725 \\ 
MFM\cite{sun2024mfm} & 33.3 & 80.0 & 80.1 & 0.948 & 0.664 & 50.8 & 83.4 & 83.4 & 0.580 & 0.722 \\ 
MMIM\cite{han2021improving} & 45.0 & 85.1 & 85.0 & 0.738 & 0.781 & 53.1 & 85.1 & 85.0 & 0.547 & 0.752 \\ 
HyCon\cite{mai2022hybrid} & 46.6 & 85.2 & 85.1 & 0.741 & 0.779 & 52.8 & 85.4 & 85.6 & 0.554 & 0.751 \\ 
UniMSE†\cite{hu2022unimse} & \underline{48.7} & 86.9 & 86.4 & 0.691 & 0.809 & \underline{54.4} & \underline{87.5} & \underline{87.5} & \underline{0.523} & 0.773 \\ 
ConFEDE†\cite{yang2023confede} & 42.3 & 85.5 & 85.5 & 0.742 & 0.782 & \textbf{54.9} & 85.8 & 85.8 & \textbf{0.522} & 0.780 \\ 

MGCL\cite{mai2023learning} & \textbf{49.3} & 86.7 & 86.7 & 0.685 & 0.707 & 53.9 & 86.4 & 86.4 & 0.535 & 0.772 \\ 
ULMD†\cite{zhu2025multimodal} & 47.8 & 85.8 & 85.7 & 0.700 & 0.799 & 53.8 & 86.0 & 85.9 & 0.531 & 0.770 \\ 

MFON† \cite{zhang2025modal} & 44.9 & 86.9 & 86.9 & 0.725 & 0.797 & 53.7 & 86.3 & 86.3 & 0.528 & 0.780 \\ 


ITHP\cite{xiao2024neuro} & 47.7 & \underline{88.5} & \underline{88.5} & \underline{0.663} & \underline{0.856} & 52.2 & 87.1 & 87.1 & 0.550 & \textbf{0.792} \\ \hline
KAN-MCP & 48.3 & \textbf{89.4} & \textbf{89.4} & \textbf{0.615} & \textbf{0.857} & 53.9 & \textbf{87.7} & \textbf{87.6} & \textbf{0.522} & \underline{0.788} \\  \hline
\end{tabular}
\vspace{-0.3cm}
\end{table*}

\begin{table}[t]
    \vspace{0.1cm}
    \caption{The comparison with baselines on the CH-SIMS v2 dataset. The baseline models are reproduced with BERT as the language model. The best results are highlighted in bold, and the runner-up results are indicated with underlines.}
    \label{tab:SIMSResult}
    \resizebox{0.48\textwidth}{!}{
        \begin{tabular}{l|c|c|c|c|c|c}
        \hline
        \multirow{2}{*}{Model} & \multicolumn{5}{c}{CH-SIMS v2} \\ \cline{2-7}
         & Acc5↑ & Acc3↑ & Acc2↑ & F1↑ & MAE↓ & Corr↑  \\ \hline
        EF-LSTM \cite{williams2018recognizing} & 53.7 & 73.5 & 80.1 & 80.0 & 0.309 & 0.700  \\ 
        LF-DNN \cite{williams2018dnn} & 51.8 & 71.2 & 77.8 & 77.9 & 0.322 & 0.668 \\
        TFN \cite{zadeh2017tensor} & 53.3 & 70.9 & 78.1 & 78.1 & 0.322 & 0.662 \\
        LMF \cite{liu2018efficient} & 51.6 & 70.0 & 77.8 & 77.8 & 0.327 & 0.651 \\
        MFN \cite{zadeh2018memory} & \underline{55.4} & 72.7 & 79.4 & 79.4 & 0.301 & 0.712 \\
        Graph-MFN \cite{zadeh2018multimodal} & 48.9 & 68.6 & 76.6 & 76.6 & 0.334 & 0.644 \\
        MulT \cite{tsai2019multimodal} & 54.6 & 74.2 & \underline{80.8} & \underline{80.7} & 0.300 & 0.738 \\
        MISA \cite{hazarika2020misa} & 47.5 & 68.9 & 78.2 & 78.3 & 0.342 & 0.671 \\
        MAG-BERT \cite{rahman2020integrating} & 49.2 & 70.6 & 77.1 & 77.1 & 0.346 & 0.641 \\
        Self-MM \cite{yu2021learning} & 53.5 & 72.7 & 78.7 & 78.6 & 0.315 & 0.691 \\
        MMIM \cite{han2021improving} & 50.5 & 70.4 & 77.8 & 77.8 & 0.339 & 0.641 \\
        AV-MC \cite{liu2022make} & 52.1 & 73.2 & 80.6 & \underline{80.7} & 0.301 & 0.721 \\
        KuDA \cite{feng2024knowledge} & 53.1 & \underline{74.3} & 80.2 & 80.1 & \underline{0.289} & \underline{0.741} \\
        
         \hline
        KAN-MCP & \textbf{57.3} & \textbf{75.0} & \textbf{81.6} & \textbf{81.7} & \textbf{0.281} & \textbf{0.742}  \\  \hline
        \end{tabular}
    }
    \vspace{-0.4cm}
\end{table}

\section{Experiments}
    We conducted extensive experiments to evaluate KAN-MCP using the CMU-MOSI \cite{zadeh2016multimodal}, CMU-MOSEI \cite{zadeh2018multimodal}, and CH-SIMS v2 \cite{liu2022make} datasets. Due to space constraints, the descriptions of the datasets, feature extraction details, evaluation metrics, training details, and baselines are provided in the Appendix. 
\subsection{Results on MSA}
    To evaluate the performance of KAN-MCP against competitive baselines, we employed two well-established MSA datasets. The results in Table \ref{tab:CMUResult} demonstrate that KAN-MCP exhibits superior performance, consistently outperforming baseline methods across most evaluation metrics on both the CMU-MOSI and CMU-MOSEI datasets. Specifically, on CMU-MOSI, KAN-MCP achieves a 0.6\% improvement in 7-class accuracy (Acc7) over ITHP \cite{xiao2024neuro}, which also employs DeBERTa \cite{he2020deberta} as its language network, along with 0.9\% enhancements in binary accuracy (Acc2) and F1 score. Notably, KAN-MCP significantly surpasses existing methods in MAE. For CMU-MOSEI, KAN-MCP outperforms the state-of-the-art ITHP algorithm by 1.7\%, 0.6\%, and 0.5\% in Acc7, Acc2, and F1 score, respectively. Furthermore, Table \ref{tab:SIMSResult} presents the robustness validation results of KAN-MCP in cross-lingual scenarios. On the Chinese dataset CH-SIMS v2, the model maintains leading performance across all metrics, including 5-class accuracy (Acc5), 3-class accuracy (Acc3), Acc2, F1 score, MAE, and correlation (Corr).


\subsection{Study of Interpretability}
    \begin{figure*}[ht]
        \vspace{-0.4cm}
        \centering
        \subfigure[CMU-MOSI]{
            \includegraphics[width=0.30\textwidth]{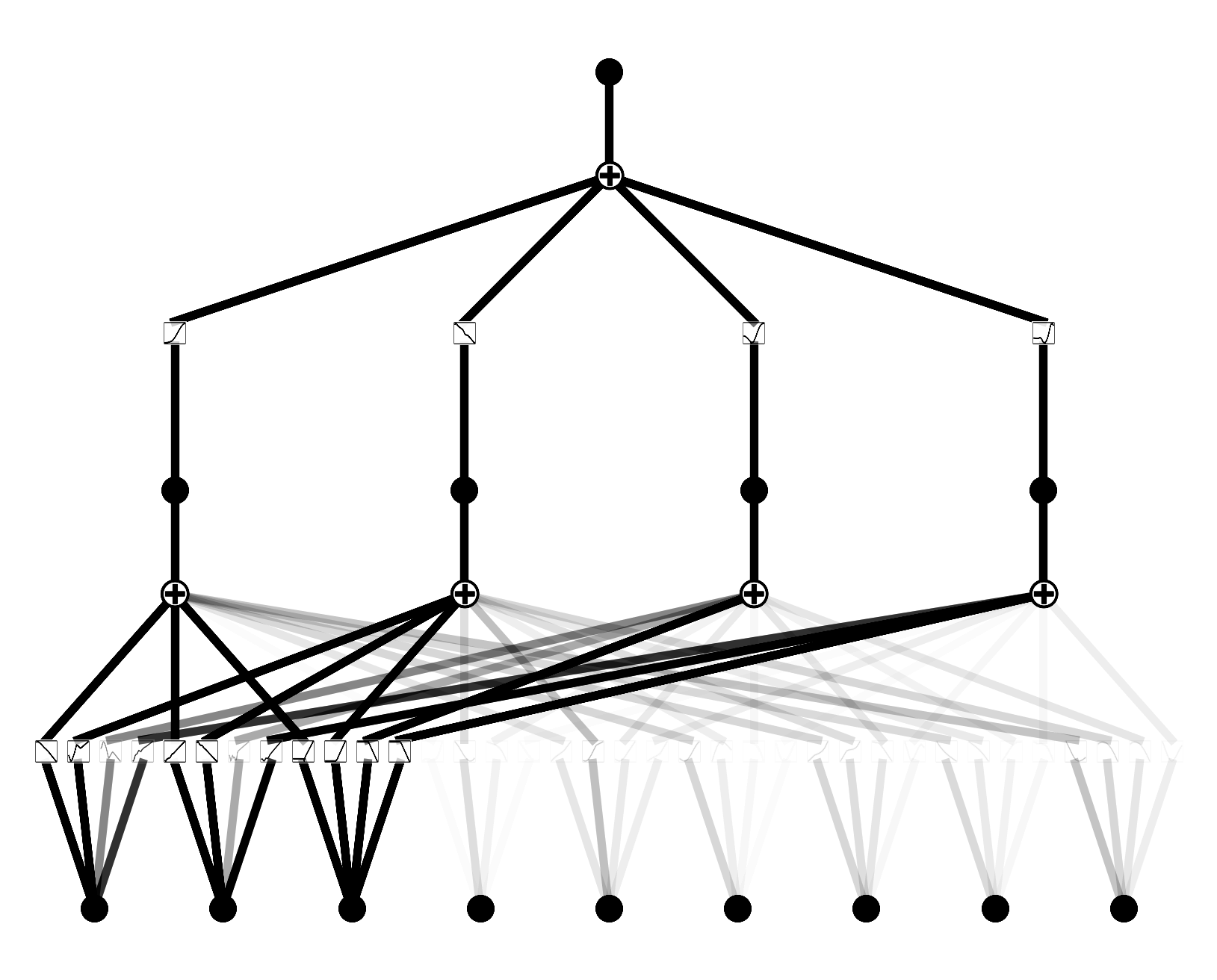}
            \label{fig:mosi}
        }
        \hfill
        \subfigure[CMU-MOSEI]{
            \includegraphics[width=0.30\textwidth]{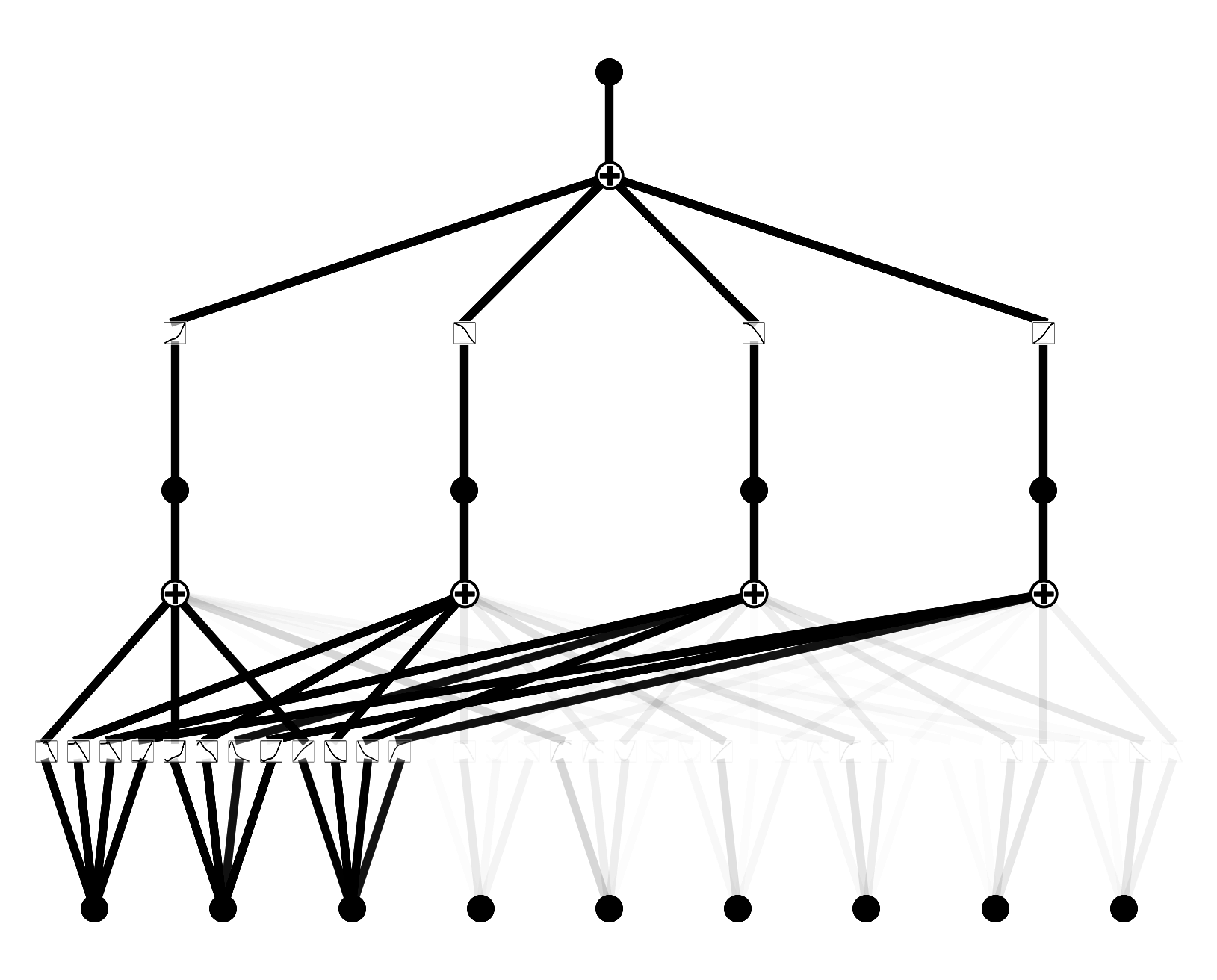}
            \label{fig:mosei}
        }
        \hfill
        \subfigure[CH-SIMSv2]{
            \includegraphics[width=0.30\textwidth]{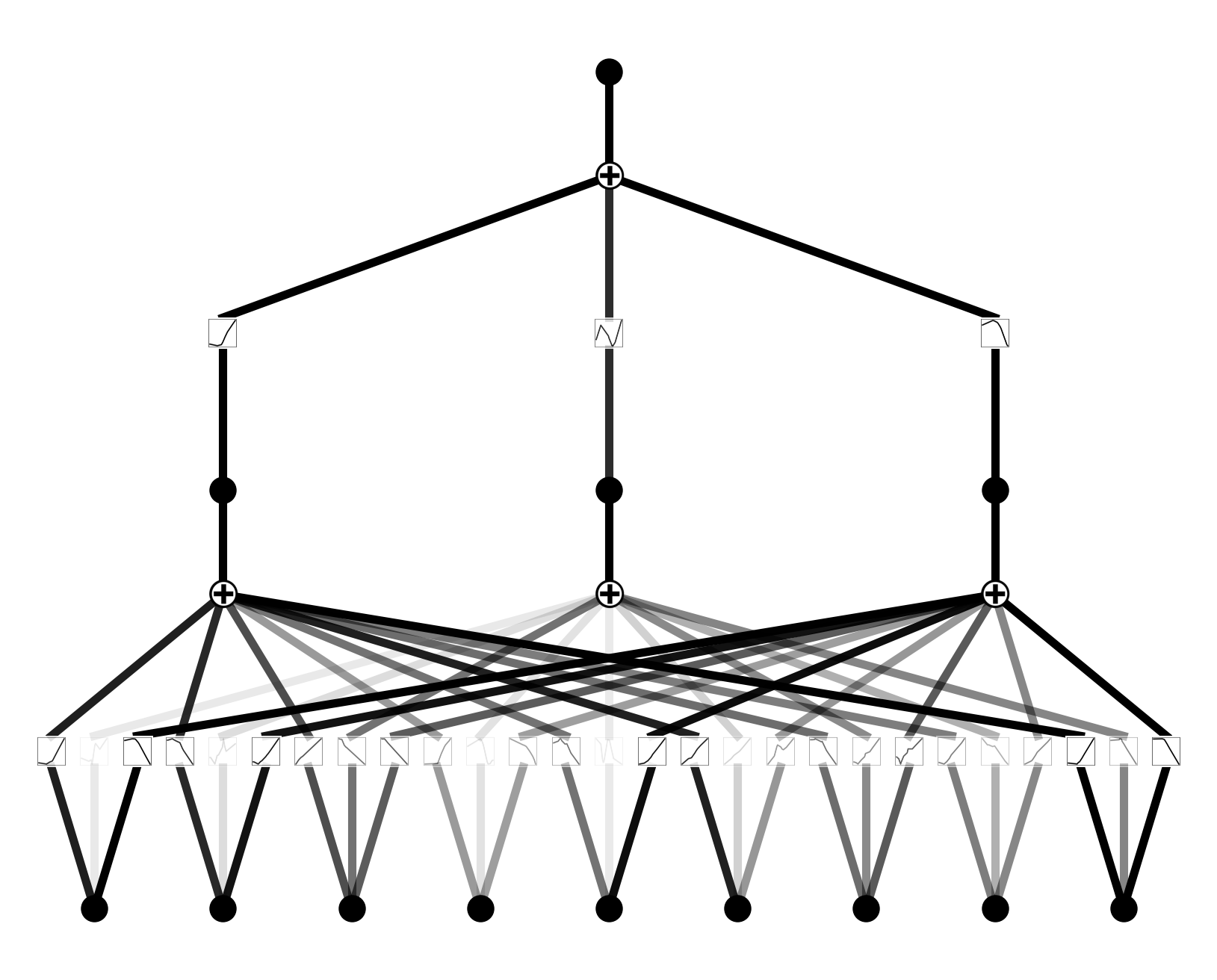}
            \label{fig:sims}
        }
        \vspace{-0.2cm}
        \caption{KAN Visualization Across Datasets: Opacity Negatively Correlates with Modality Contribution Weights.}
        \label{fig:visualizaton}
        \vspace{-0.1cm}
    \end{figure*}
    
    In this section, we use the visualization capabilities of the KAN model to analyze the multimodal fusion prediction process, with results illustrated in Figure \ref{fig:visualizaton}. By observing connection transparency, we can assess the contribution of different modalities. For the CMU-MOSI and CMU-MOSEI datasets, Figures \ref{fig:mosi} and \ref{fig:mosei} reveal that the text modality exhibits significantly lower connection transparency between data points and intermediate nodes compared to other modalities. This observation indicates that textual information plays a fundamental and critical role in the model, which aligns with existing research conclusions \cite{mai2021analyzing,mai2019divide} and further validates the pivotal status of textual information in sentiment analysis tasks.

    In contrast, the audio and visual modalities show higher connection transparency. This suggests that audio and visual information are less directly impactful. However, this does not mean that the roles of the audio and visual modalities can be neglected. On the contrary, they primarily serve indispensable supplementary functions during model operation. In specific scenarios, audio and visual information provide cues to assist the model. For instance, in complex multimodal emotional scenarios, textual information alone might fail to capture all affective clues, whereas audio and visual signals could offer multi-dimensional supplementary emotional hints to enhance comprehensive understanding of emotional expressions. This complementary effect not only enriches the model's feature inputs but also optimizes its performance, enabling more flexible and accurate sentiment analysis and prediction in diverse real-world applications.

    For the CH-SIMS v2 dataset, Figure \ref{fig:sims} demonstrates relatively more balanced connection intensities between textual, audio, and visual data points and intermediate nodes, without a clearly dominant modality. This phenomenon stems from the comparable sentiment analysis capabilities of all three modalities in the CH-SIMS v2 dataset, a finding that is further corroborated in subsequent analyses of imbalanced multimodal learning.

    Furthermore, KAN can provide more intuitive mathematical expressions and enhanced interpretability through symbolic manipulation. Detailed information about the experimental results is presented in the Appendix.



\subsection{Imbalanced Multimodal Learning Analysis}
    
    \begin{figure}[t]
        \centering
        \subfigure[CMU-MOSI Text]{
            \includegraphics[width=0.14\textwidth]{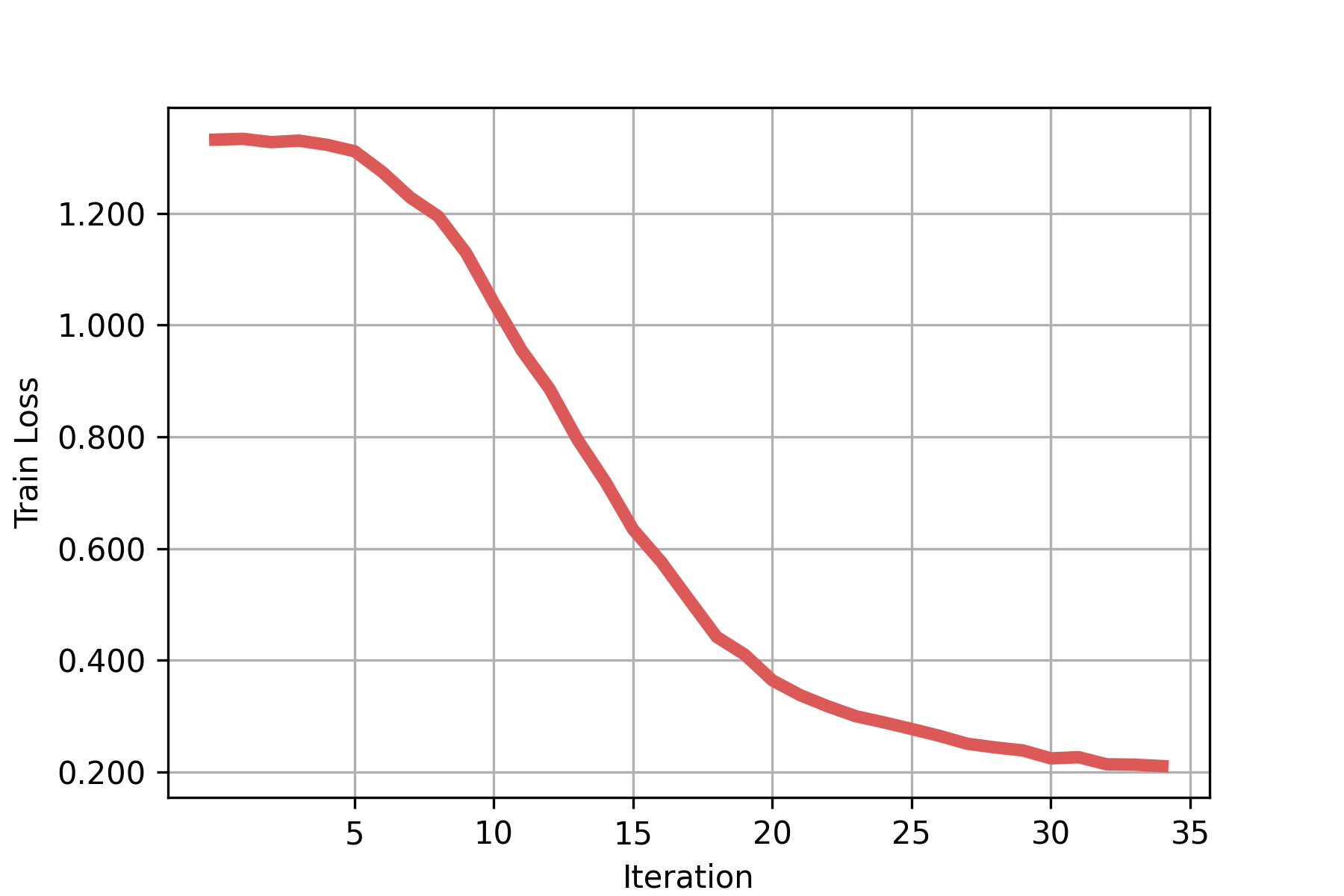}
        }
        \subfigure[CMU-MOSI Audio]{
            \includegraphics[width=0.14\textwidth]{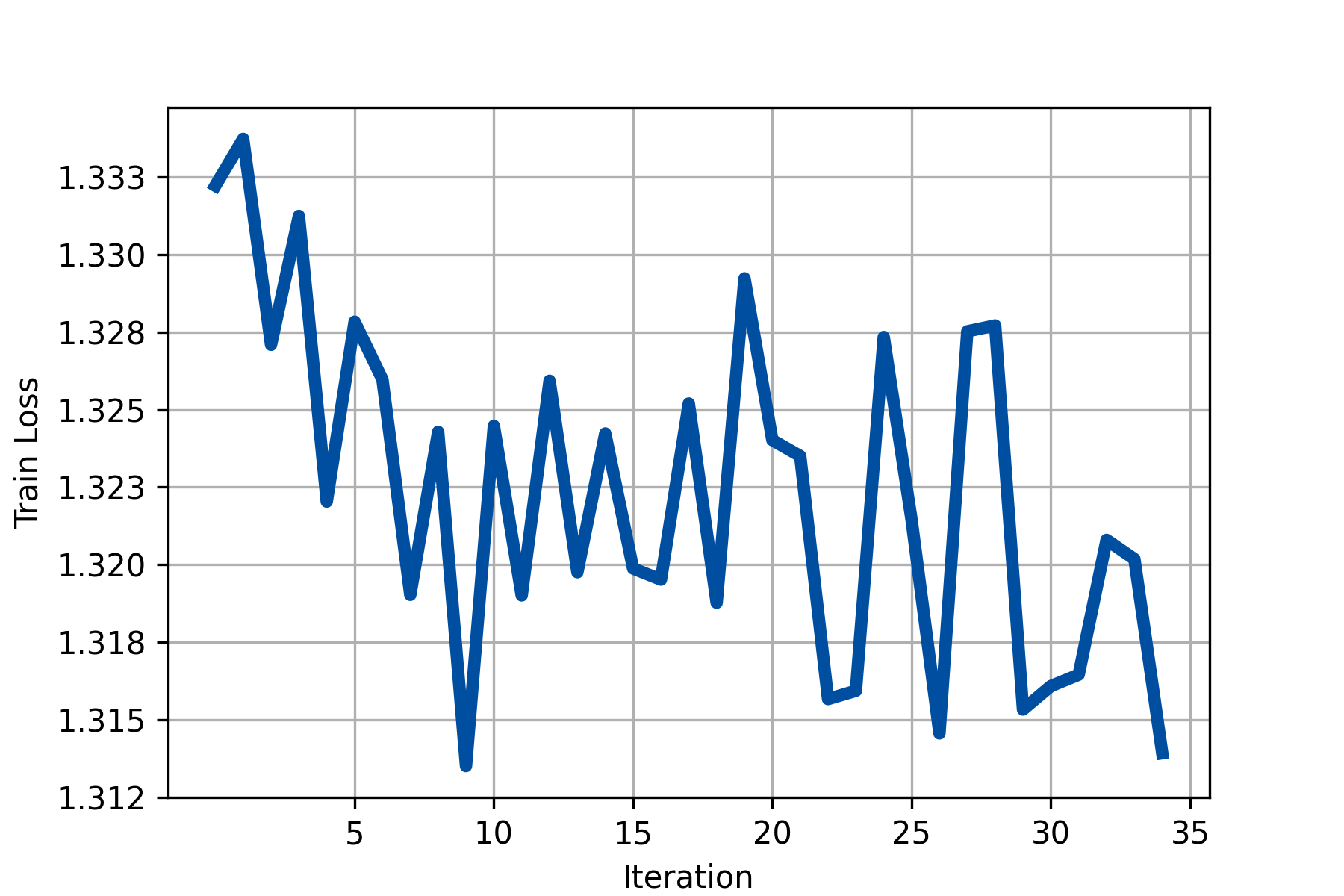}
        }
        \subfigure[CMU-MOSI Visual]{
            \includegraphics[width=0.14\textwidth]{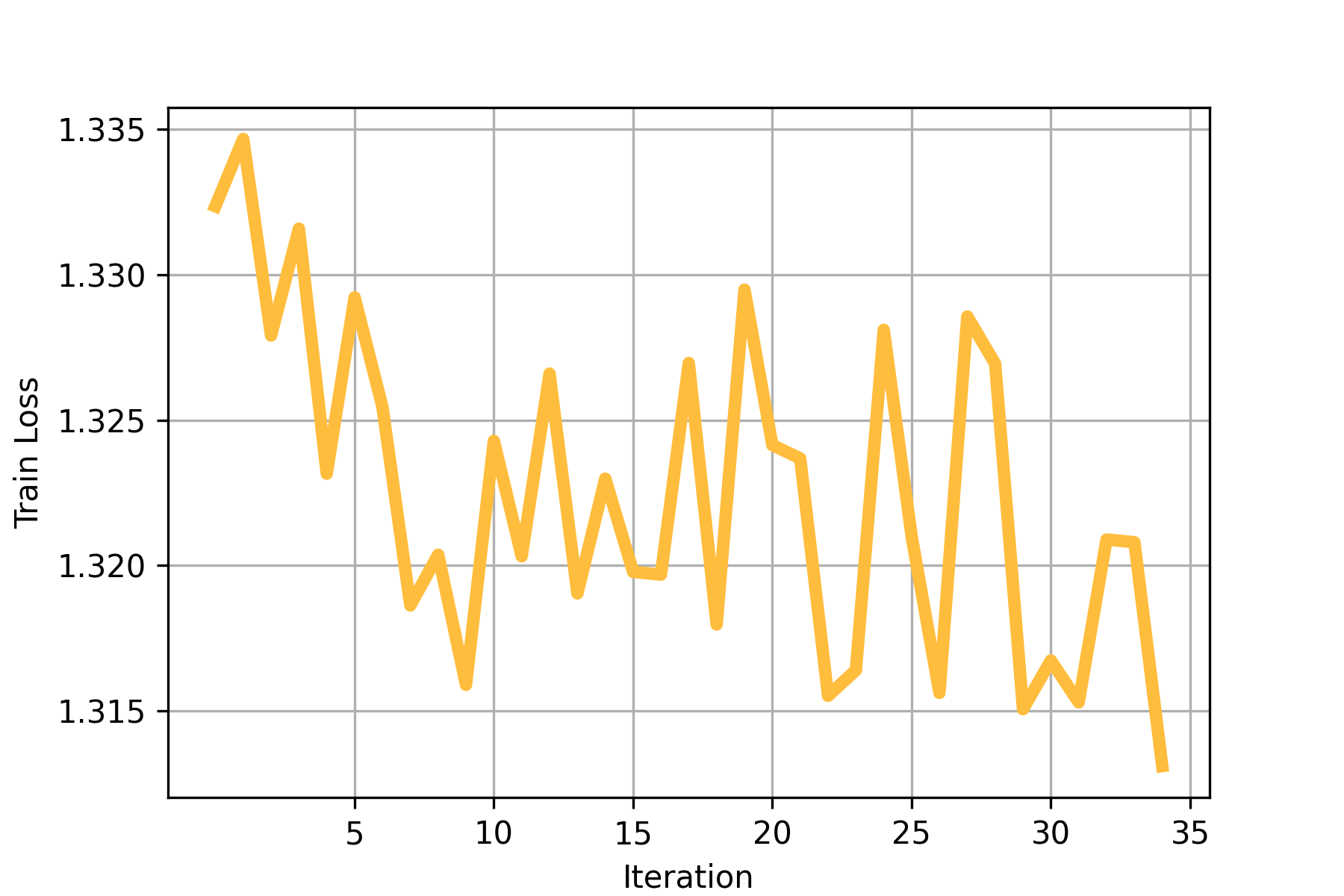}
        }
        \subfigure[CMU-MOSEI Text]{
            \includegraphics[width=0.14\textwidth]{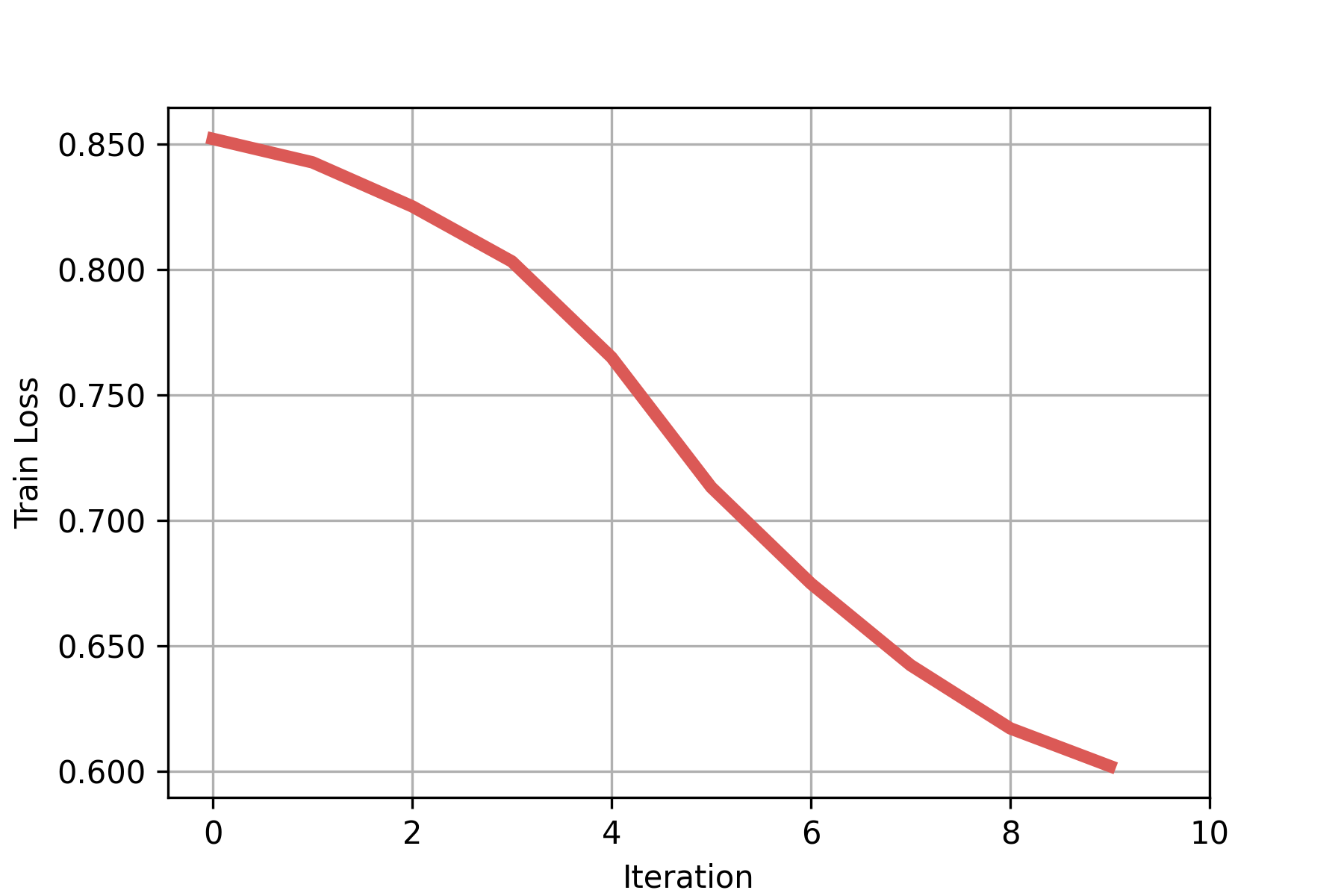}
        }
        \subfigure[CMU-MOSEI Audio]{
            \includegraphics[width=0.14\textwidth]{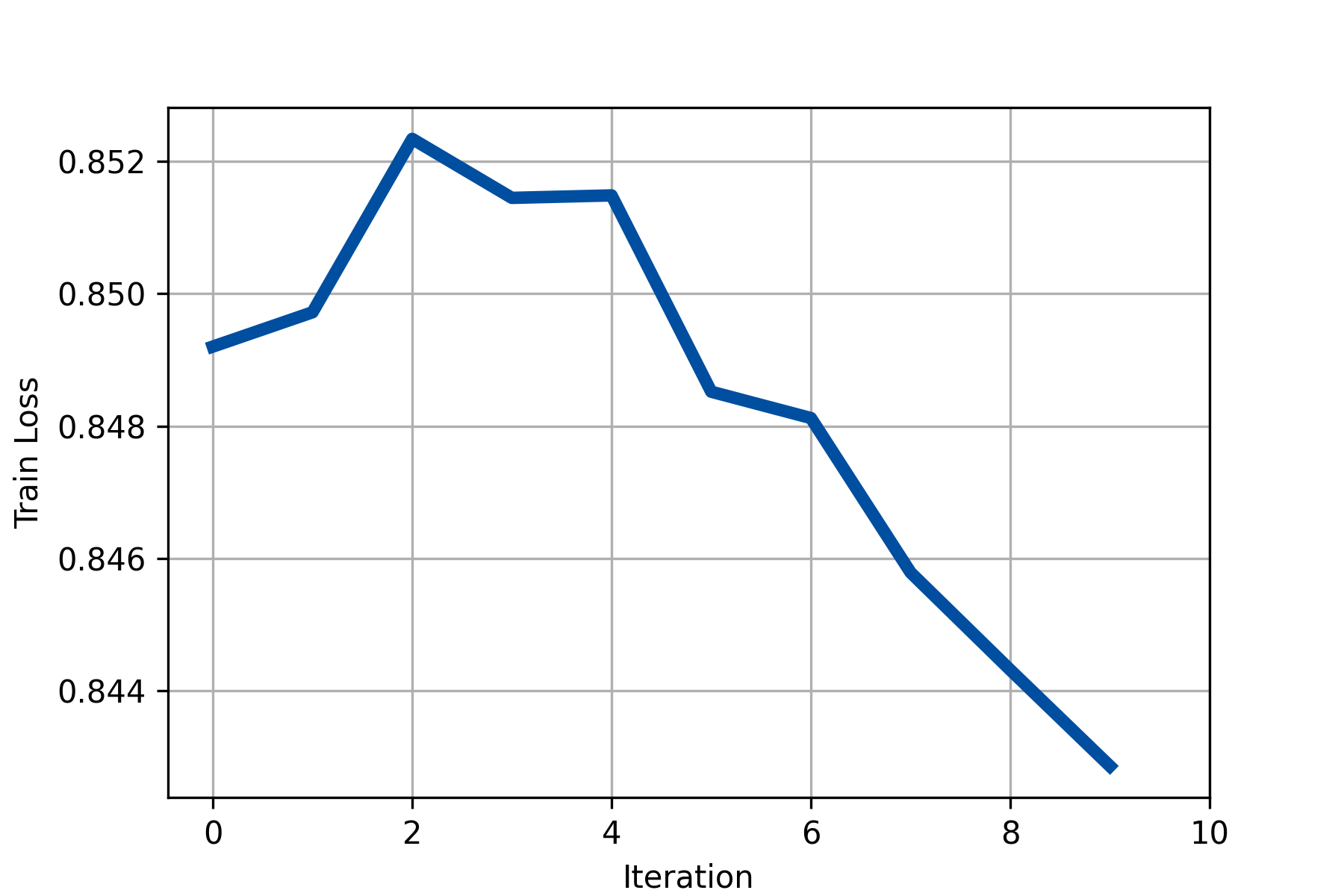}
        }
        \subfigure[CMU-MOSEI Visual]{
            \includegraphics[width=0.14\textwidth]{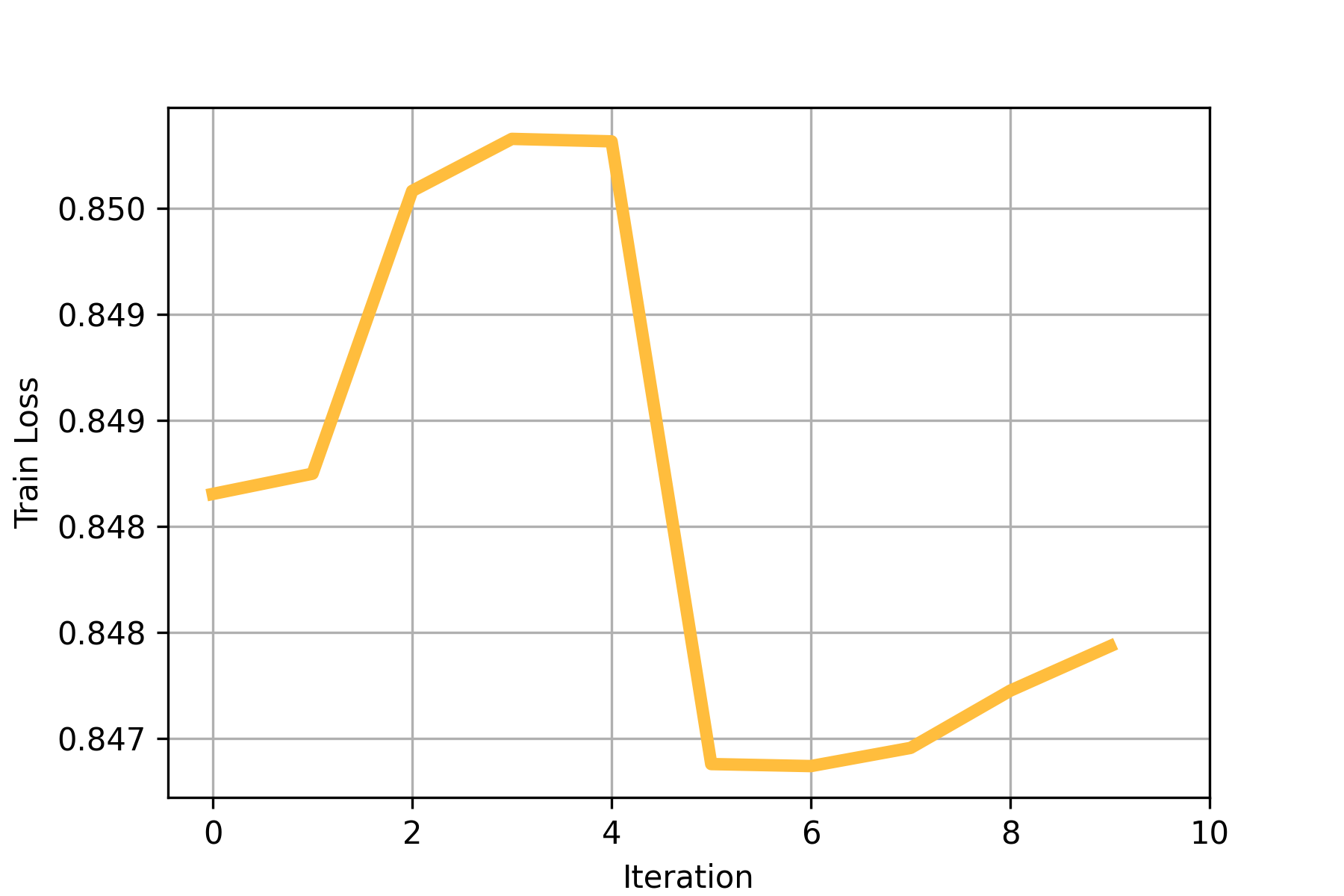}
        }
        \subfigure[SIMSv2 Text]{
            \includegraphics[width=0.14\textwidth]{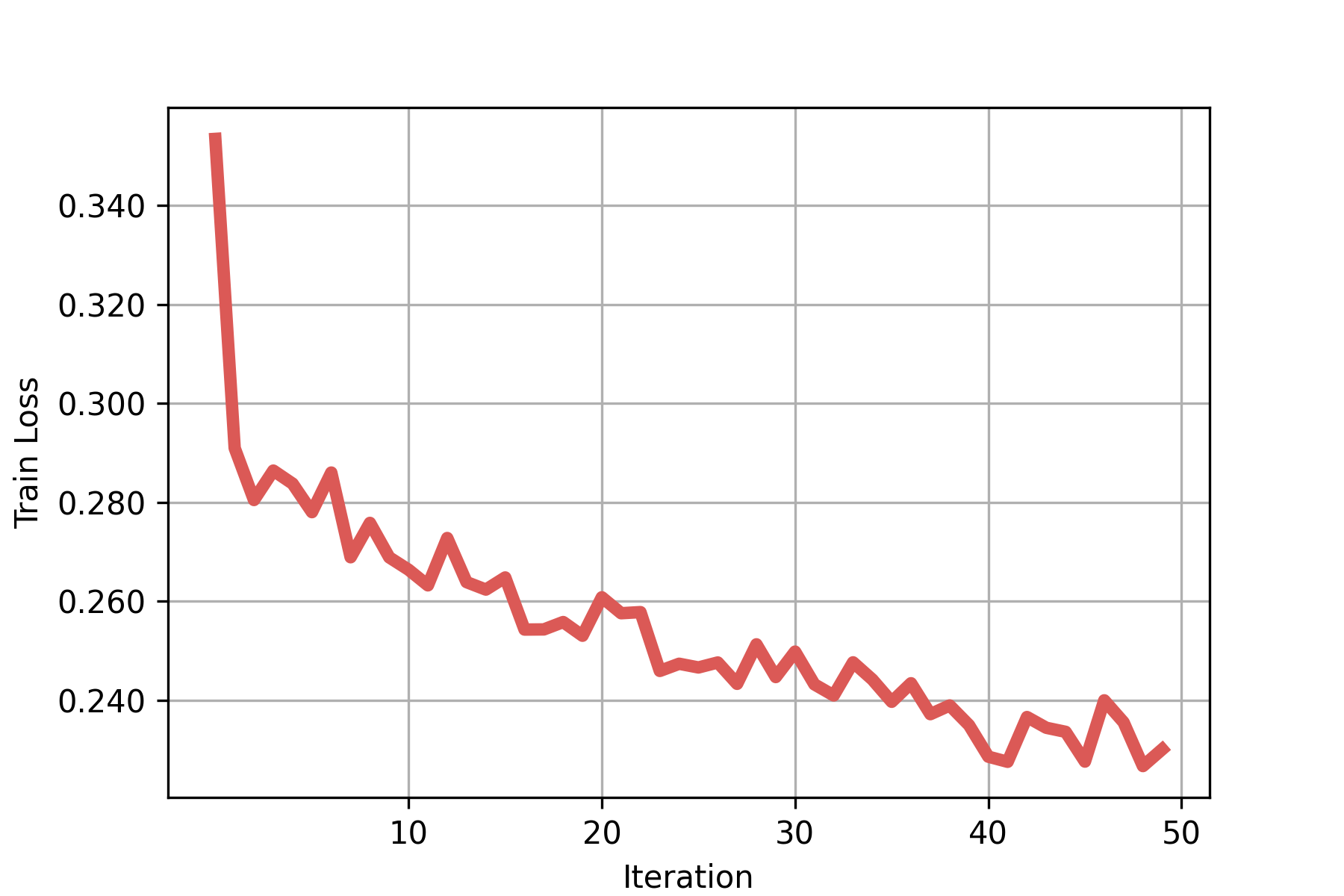}
        }
        \subfigure[SIMSv2 Audio]{
            \includegraphics[width=0.14\textwidth]{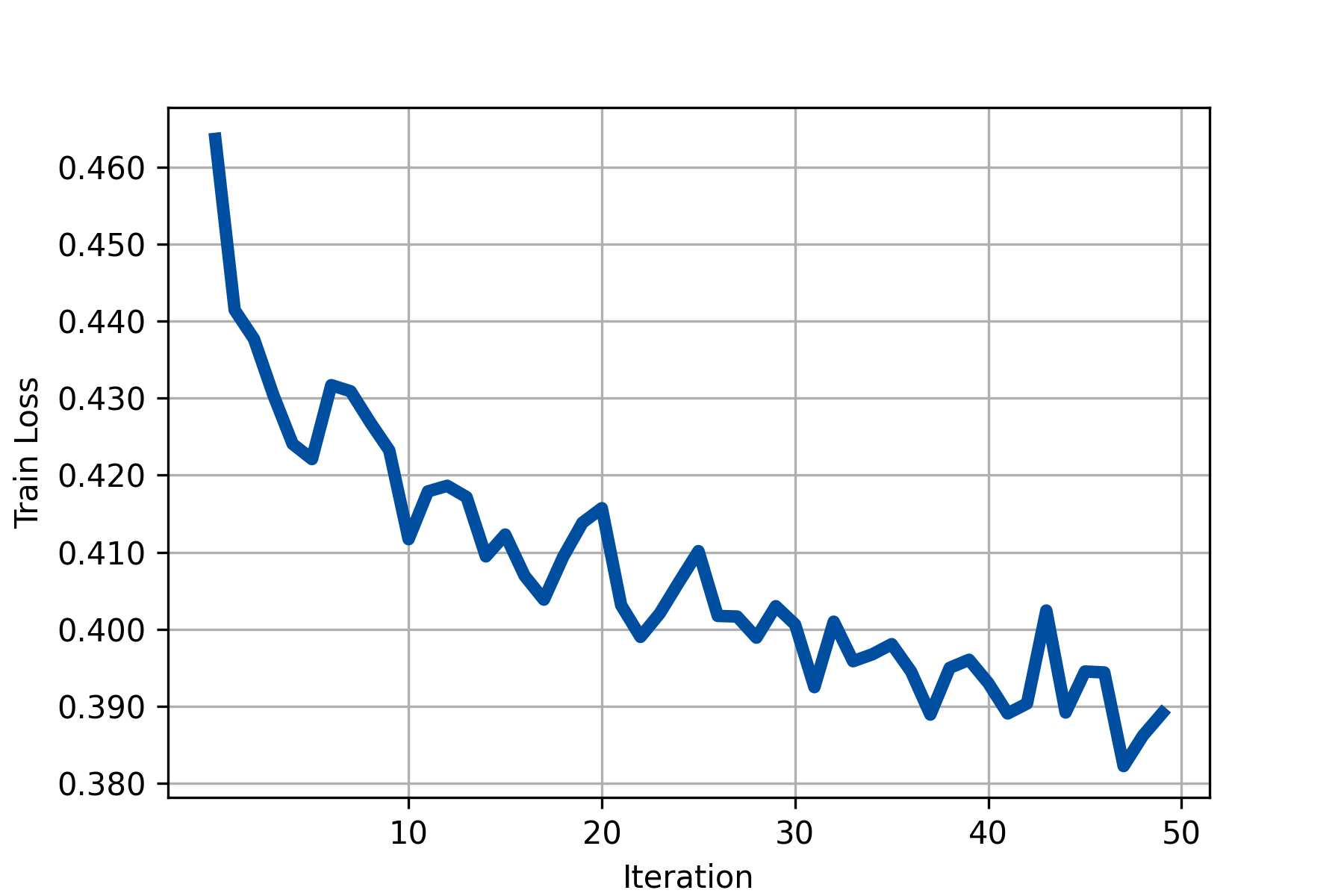}
        }
        \subfigure[SIMSv2 Visual]{
            \includegraphics[width=0.14\textwidth]{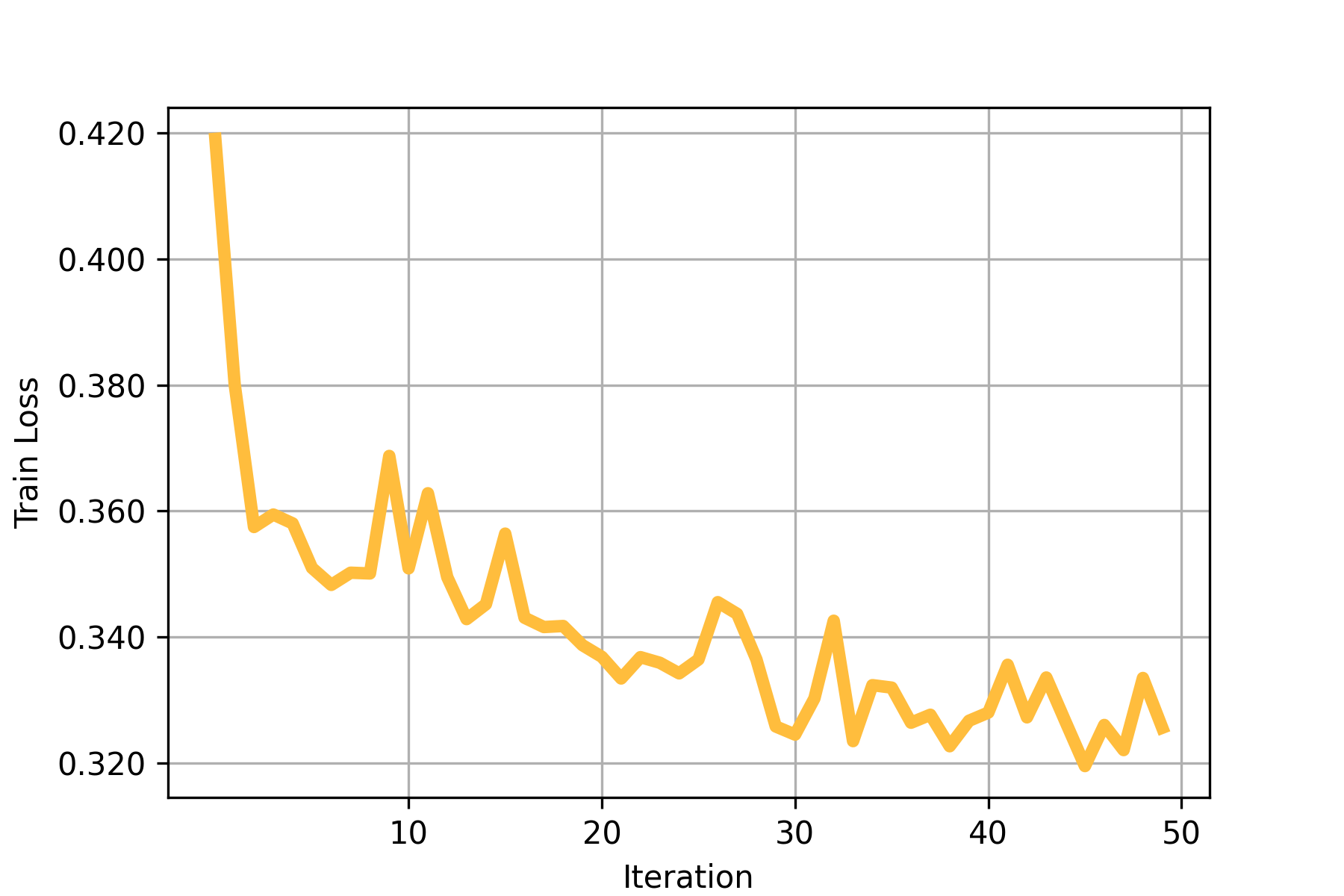}
        }
        \caption{The Learning Curves of the Training Losses for Audio, Visual, and Text Modalities.}
        \label{fig:loss curves}
        \vspace{-0.5cm}
    \end{figure}
    
    In this section, we present the learning curves of unimodal losses across three datasets. As illustrated in Figure \ref{fig:loss curves}, the training losses of the textual modality  demonstrate a continuous decline across all three datasets, indicating the model's pronounced advantages in textual feature extraction and sentiment semantic comprehension. Notably, the audio and visual modalities in CMU-MOSI and CMU-MOSEI datasets exhibit significant training instability, characterized by marked fluctuations in their loss curves. This phenomenon aligns with the imbalanced modality contribution patterns revealed in Figure \ref{fig:visualizaton}. In contrast, the multimodal training process of CH-SIMS V2 dataset displays superior synergy, with non-textual modalities (audio, visual) demonstrating smooth descending loss curves whose convergence rates remain synchronized with the textual modality. These disparities in training stability effectively elucidate the observed differences in modality contributions across datasets for MSA tasks: When non-textual modalities achieve sufficient learning effectiveness (as in CH-SIMS V2), the model attains more balanced multimodal representation fusion; conversely, it may develop over-reliance on dominant modalities (e.g., textual modality in CMU datasets) when such effectiveness is lacking. This finding corroborates the aforementioned characteristics of modality contribution equilibrium.

    The scarcity of semantic information in non-textual modalities within the CMU-MOSI and CMU-MOSEI datasets may stem from two primary factors. First, acoustic signals and visual information inherently contain noise interference, and the instability in raw data quality directly impacts the robustness of the feature extraction process. Second, the feature extraction architectures employed in these datasets (e.g., Facet and COVAREP) may inadequately capture discriminative features pertinent to sentiment analysis tasks.

    From the above analysis, it is evident that the quality of modal representations exhibits a significant correlation with the transparency of connecting lines in KAN visualization diagrams. By observing the transparency of connection lines between modal data points and intermediate nodes, the quality of modal representations can be indirectly assessed. For instance, high-quality modalities typically manifest as connection lines with lower transparency (more opaque), whereas low-quality modalities are characterized by lines with higher transparency (more translucent). This visual analytical approach provides an intuitive tool for understanding the relationship between modal representation quality and model dependency, while also offering valuable insights for optimizing the design of multimodal models.


\subsection{Ablation Study}    

    \begin{table}[t]
    \caption{Ablation experiments on CMU-MOSEI dataset.}
    \centering
    \resizebox{0.47\textwidth}{!}{
        \begin{tabular}{l|c|c|c|c|c}
            \hline
            \textbf{} & Acc7(↑) & Acc2(↑) & F1(↑) & MAE(↓) & Corr(↑) \\
            \hline
            w/o DRD-MIB & 47.3 & 82.6 & 82.2 & 0.614 & 0.703  \\
            w/o KAN & 52.9 & 86.7 & 86.6 & 0.535 & 0.780  \\
            w/o MMPareto & 53.1 & 86.5 & 86.4 & 0.536 & 0.782 \\ \hline
            KAN-MCP & \textbf{53.9} & \textbf{87.7} & \textbf{87.6} & \textbf{0.522} & \textbf{0.788} \\
            \hline
        \end{tabular}
    }
    \label{tab:ablation}
    \end{table}
    
    \begin{table}[t]
        \caption{Comparative Analysis of MMPareto's Effectiveness.}
        \label{tab:MMPareto Effectiveness}
        \centering
        \newcolumntype{C}[1]{>{\centering\arraybackslash}p{#1}}
        \resizebox{0.47\textwidth}{!}{
            \begin{tabular}{c|C{2cm}|c|c|c|c}
            \hline
            Dataset & MMPareto & Acc2 & F1 & MAE & Corr \\
            \hline
            \multirow{2}{*}{CMU-MOSI} & w/o & 87.3 & 87.3 & 0.647 & 0.839 \\
             & w & 89.4 & 89.4 & 0.615 & 0.857 \\
            \hline
            \multirow{2}{*}{CMU-MOSEI} & w/o & 86.5 & 86.4 & 0.536 & 0.782 \\
             & w & 87.7 & 87.6 & 0.522 & 0.788 \\
            \hline
            \multirow{2}{*}{CH-SIMSv2} & w/o & 81.2 & 81.3 & 0.282 & 0.743 \\
             & w & 81.6 & 81.7 & 0.281 & 0.742 \\
            \hline
            \end{tabular}
        }
        \vspace{-0.2cm}
    \end{table}

    \begin{table}[!t]
        \caption{Comparison and Performance Analysis of DRD-MIB and Common Dimensionality Reduction Methods (Autoencoder, Transformer, Fully Connected Layer) on the CMU-MOSEI Dataset Under MMPareto Conditions.}
        \label{tab:DRD-MIB-compare}
        \centering
        \newcolumntype{C}[1]{>{\centering\arraybackslash}p{#1}}
        \resizebox{0.47\textwidth}{!}{
            \begin{tabular}{c|C{2cm}|c|c|c|c|c}
            \hline
             & MMPareto & Acc7 & Acc2 & F1 & MAE & Corr \\
             \hline
            \multirow{2}{*}{FC} & w/o & 53.2 & 86.4 & 86.3 & 0.527 & 0.777 \\
             & w & 52.0 & 85.4 & 85.4 & 0.557 & 0.753 \\
            \hline
            \multirow{2}{*}{AE} & w/o & 53.0 & 86.4 & 86.4 & 0.532 & 0.779 \\
             & w & 47.3 & 82.6 & 82.2 & 0.614 & 0.703 \\
            \hline
            \multirow{2}{*}{Transformer} & w/o & 53.0 & 86.4 & 86.3 & 0.542 & 0.779 \\
             & w & 50.9 & 85.0 & 85.0 & 0.570 & 0.768 \\
            \hline
            \multirow{2}{*}{DRD-MIB} & w/o & 53.1 & 86.5 & 86.4 & 0.536 & 0.782 \\
             & w & 53.9 & 87.7 & 87.6 & 0.522 & 0.788 \\
            \hline
            \end{tabular}
        }
    \end{table}

    \begin{figure}[t]
        \vspace{-0.4cm}
        \centering
        \subfigure[DRD-MIB]{
            \includegraphics[width=0.21\textwidth]{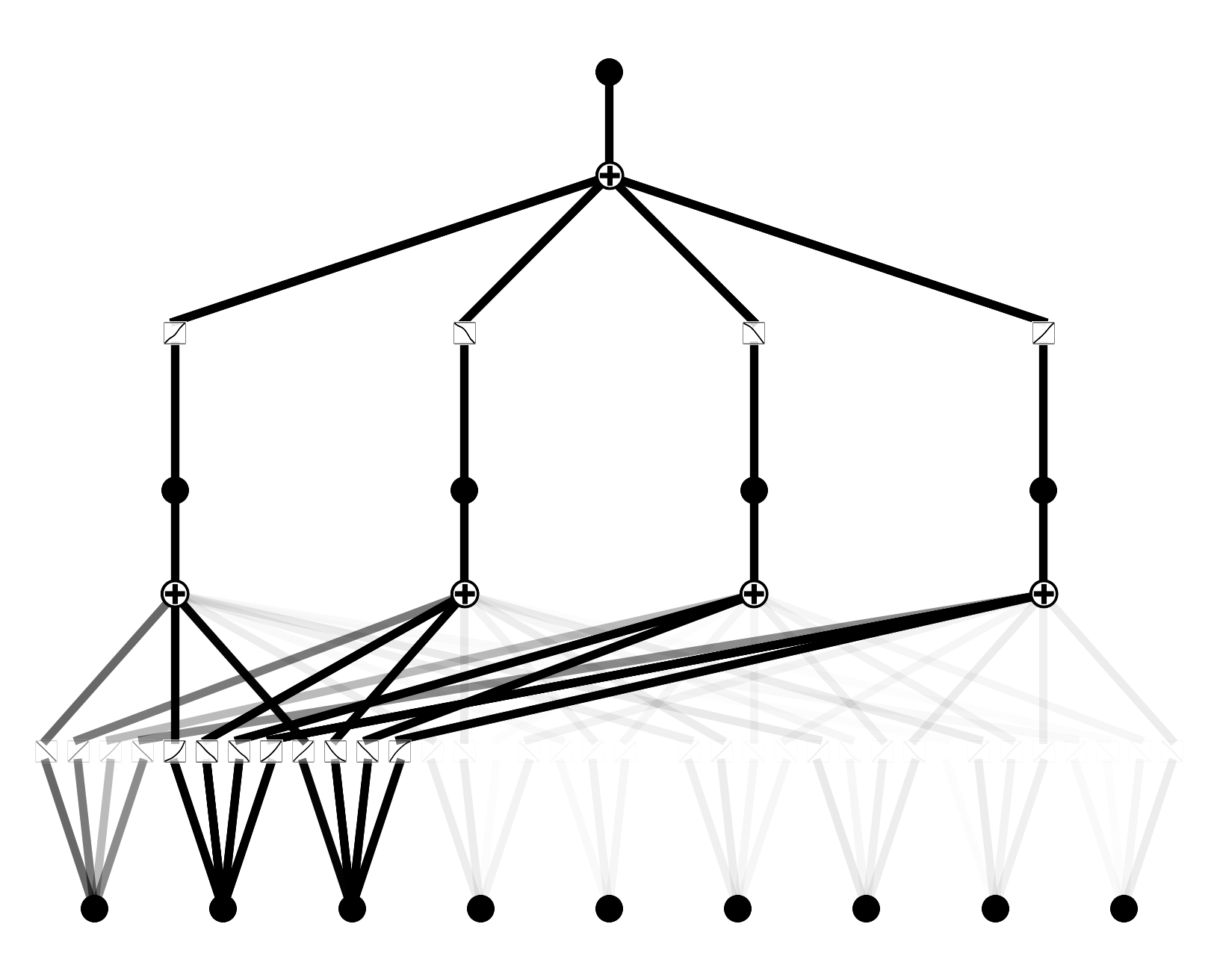}
            \label{fig:mosei_DRD-MIB}
        }
        \subfigure[AE]{
            \includegraphics[width=0.21\textwidth]{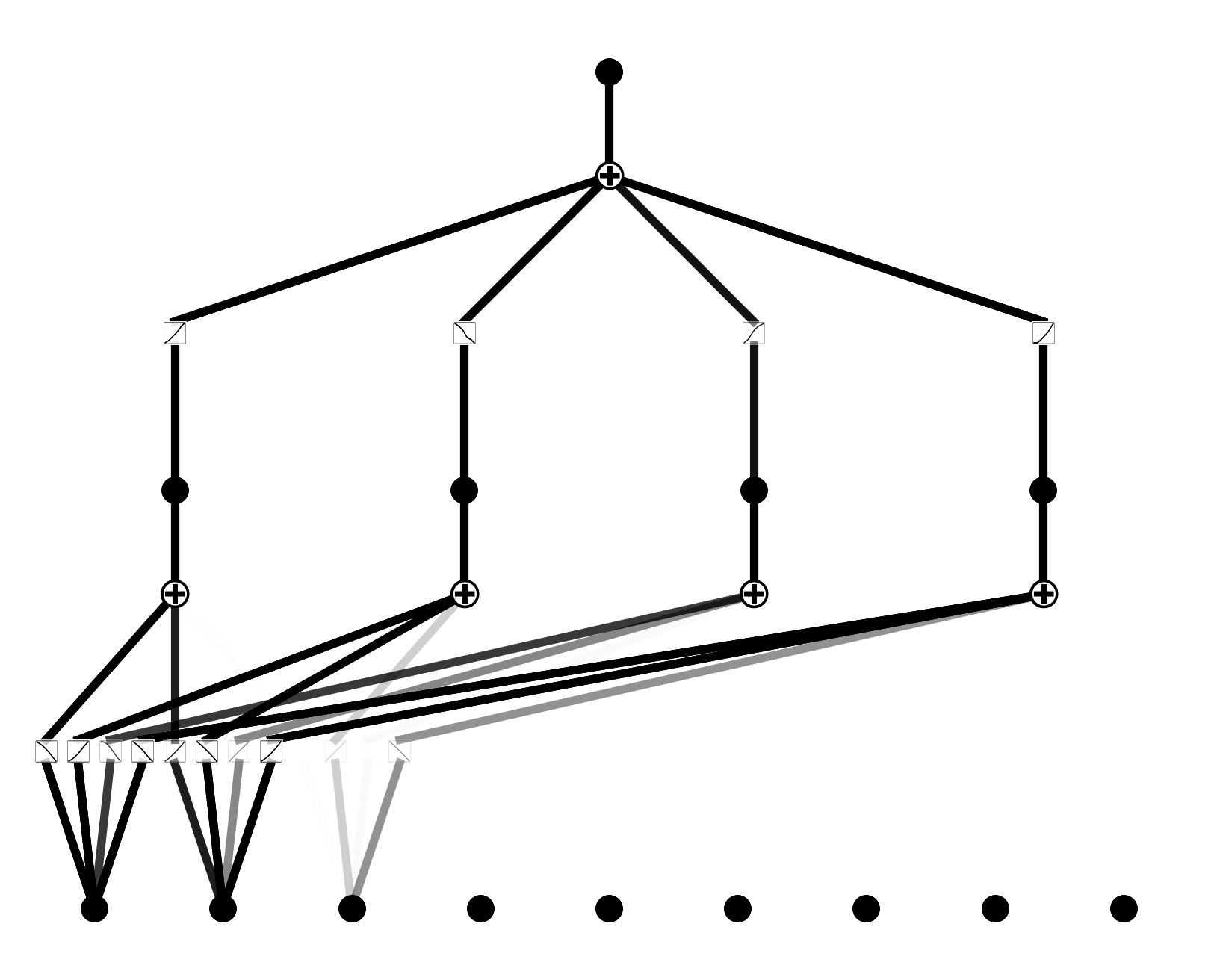}
            \label{fig:mosei_AE}
        }
        \subfigure[Transformer]{
            \includegraphics[width=0.21\textwidth]{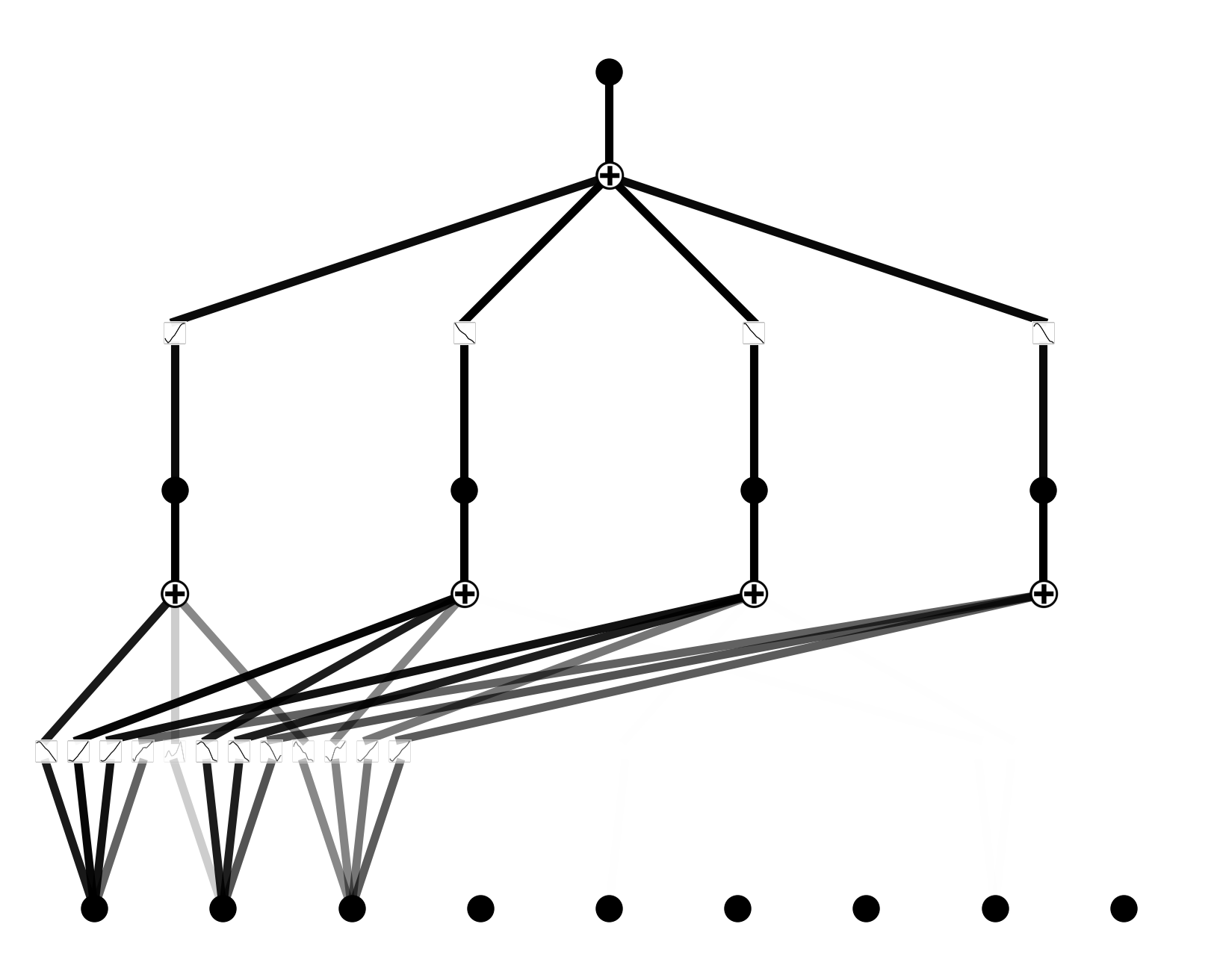}
            \label{fig:mosei_tf}
        }
        \subfigure[FC]{
            \includegraphics[width=0.21\textwidth]{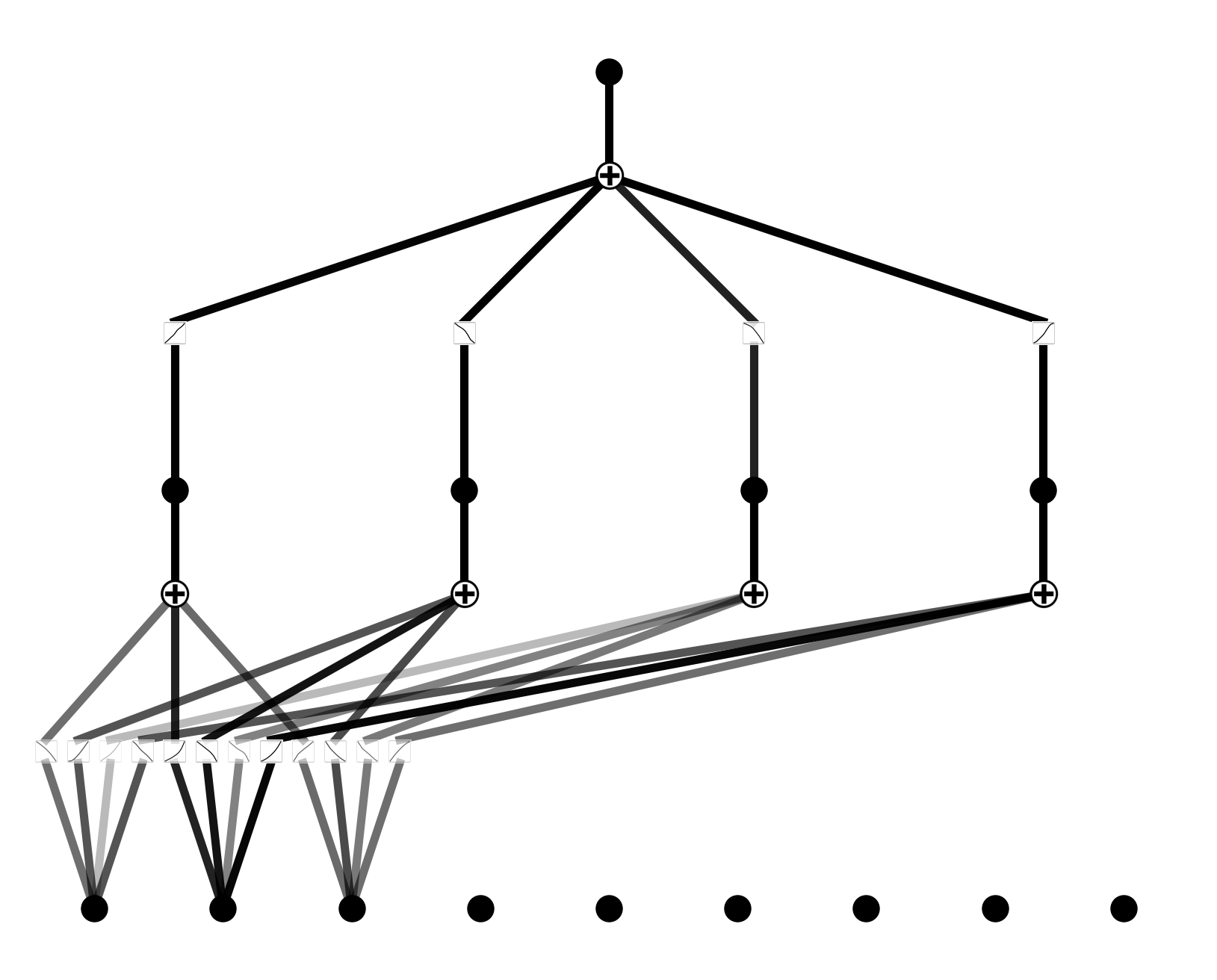}
            \label{fig:mosei_fc}
        }
        \caption{KAN Visualization of DRD-MIB and Common Dimensionality Reduction Methods (AutoEncoder, Transformer, Fully Connected Layer) on the CMU-MOSEI Dataset Under MMPareto-Free Conditions.}
        \label{fig:DRD-MIBandAEVisual}
        \vspace{-0.2cm}
    \end{figure}

    In this section, we conducted extensive ablation studies to evaluate the effectiveness of each component in KAN-MCP:
    
    \subsubsection{The importance of KAN} In the "W/O KAN" setting, we replaced KAN with a a multilayer perceptron (MLP).  As shown in Table \ref{tab:ablation}, removing KAN caused a 0.4\% drop in Acc7, a more significant 1.2\% decrease in Acc2, and a 1.1\% reduction in F1 score. These results confirm that KAN enhances interpretability while significantly improving accuracy in multimodal fusion prediction.

    \subsubsection{Discussion on DRD-MIB} In the "W/O DRD-MIB" setting, as shown in Table \ref{tab:ablation}, the performance of KAN-MCP exhibited significant degradation when an autoencoder (AE) was used as the feature dimensionality reduction component. To further investigate this phenomenon, we removed the MMPareto algorithm and compared DRD-MIB with conventional dimensionality reduction methods, including AE, Transformer, and fully connected layers (FC). The results in Table \ref{tab:DRD-MIB-compare} indicate that, while conventional methods achieved comparable overall performance to DRD-MIB, an in-depth analysis of KAN’s visualization results, as shown in Figure \ref{fig:mosei_AE},\ref{fig:mosei_tf}, and \ref{fig:mosei_fc}, revealed a critical limitation: traditional methods predominantly captured textual modality information while neglecting audio and visual modality features. This highlights the inherent inability of conventional dimensionality reduction methods to effectively integrate multimodal information, particularly in scenarios where certain modalities contain limited information content.

    In contrast, when DRD-MIB was employed as the dimensionality reduction component, as shown in Figure \ref{fig:mosei_DRD-MIB}, the model successfully learned textual modality features while simultaneously capturing audio and visual modality information, thereby achieving comprehensive and effective multimodal integration. These results further validate the superiority of DRD-MIB in feature dimensionality reduction. Specifically, DRD-MIB leverages its unique mechanism to learn more balanced and high-quality unimodal representations, thereby significantly enhancing model performance in multimodal tasks. This finding underscores the distinct advantages of DRD-MIB in multimodal feature dimensionality reduction and information integration.

    \subsubsection{Discussion on MMPareto} In the "W/O MMPareto" setting, we analyzed performance across three datasets. As shown in Table \ref{tab:MMPareto Effectiveness}, performance declined across all datasets, with notable degradation observed on CMU-MOSI and CMU-MOSEI, where Acc2 and F1 scores decreased by 2.1\% and 1.2\%, respectively. In contrast, CH-SIMS v2 exhibited only marginal declines (0.4\% in Acc2 and F1). These results suggest that MMPareto effectively leverages auxiliary unimodal information to enhance multimodal learning performance when significant inter-modal information disparities exist and weaker modalities contain non-harmful yet useful information. However, its performance gains diminish when modalities exhibit relatively balanced information content.

    Further comparative experiments revealed compatibility issues between traditional dimensionality reduction methods (AE, Transformer, FC) and the MMPareto optimizer. As shown in Table \ref{tab:DRD-MIB-compare}, unexpected performance degradation occurred when conventional dimensionality reduction components were used. We hypothesize that this stems from the inability of traditional methods to effectively filter redundant features, leading to increased gradient conflicts between unimodal and multimodal loss directions, thereby impairing model performance. This suggests that the efficacy of MMPareto may be constrained when weak modality features suffer from information loss.

    In summary, our proposed MCPareto framework integrates the DRD-MIB denoising dimensionality reduction method and the MMPareto optimization algorithm. This combination suppresses noise interference while enhancing the contribution weights of secondary modalities, thereby improving model generalization. This two-stage optimization strategy not only addresses modality imbalance but also significantly enhances robustness in complex multimodal scenarios, offering a more reliable solution for multimodal learning.

\section{Conclusion}
    This paper proposes a novel MSA framework, KAN-MCP, designed to address the interpretability of multimodal fusion and modality imbalance issues. By integrating KAN with MCPareto, the framework achieves transparent cross-modal interaction modeling and balanced modality learning. Experimental results demonstrate that KAN-MCP significantly outperforms existing methods on CMU-MOSI, CMU-MOSEI, and CH-SIMS v2 datasets. Furthermore, KAN's visualization module intuitively reveals the contributions of different modalities in sentiment judgment, validating the dominance of textual modality in CMU datasets and the synergistic balance effect among modalities in the CH-SIMS v2 dataset.

\bibliographystyle{ACM-Reference-Format}
\bibliography{references}


\begin{thebibliography}{71}


\ifx \showCODEN    \undefined \def \showCODEN     #1{\unskip}     \fi
\ifx \showDOI      \undefined \def \showDOI       #1{#1}\fi
\ifx \showISBNx    \undefined \def \showISBNx     #1{\unskip}     \fi
\ifx \showISBNxiii \undefined \def \showISBNxiii  #1{\unskip}     \fi
\ifx \showISSN     \undefined \def \showISSN      #1{\unskip}     \fi
\ifx \showLCCN     \undefined \def \showLCCN      #1{\unskip}     \fi
\ifx \shownote     \undefined \def \shownote      #1{#1}          \fi
\ifx \showarticletitle \undefined \def \showarticletitle #1{#1}   \fi
\ifx \showURL      \undefined \def \showURL       {\relax}        \fi
\providecommand\bibfield[2]{#2}
\providecommand\bibinfo[2]{#2}
\providecommand\natexlab[1]{#1}
\providecommand\showeprint[2][]{arXiv:#2}

\bibitem[Adadi and Berrada(2018)]%
        {adadi2018peeking}
\bibfield{author}{\bibinfo{person}{Amina Adadi} {and} \bibinfo{person}{Mohammed Berrada}.} \bibinfo{year}{2018}\natexlab{}.
\newblock \showarticletitle{Peeking inside the black-box: a survey on explainable artificial intelligence (XAI)}.
\newblock \bibinfo{journal}{\emph{IEEE access}}  \bibinfo{volume}{6} (\bibinfo{year}{2018}), \bibinfo{pages}{52138--52160}.
\newblock


\bibitem[Aghaeipoor et~al\mbox{.}(2023)]%
        {aghaeipoor2023fuzzy}
\bibfield{author}{\bibinfo{person}{Fatemeh Aghaeipoor}, \bibinfo{person}{Mohammad Sabokrou}, {and} \bibinfo{person}{Alberto Fern{\'a}ndez}.} \bibinfo{year}{2023}\natexlab{}.
\newblock \showarticletitle{Fuzzy rule-based explainer systems for deep neural networks: From local explainability to global understanding}.
\newblock \bibinfo{journal}{\emph{IEEE Transactions on Fuzzy Systems}} \bibinfo{volume}{31}, \bibinfo{number}{9} (\bibinfo{year}{2023}), \bibinfo{pages}{3069--3080}.
\newblock


\bibitem[Alemi et~al\mbox{.}(2016)]%
        {alemi2016deep}
\bibfield{author}{\bibinfo{person}{Alexander~A Alemi}, \bibinfo{person}{Ian Fischer}, \bibinfo{person}{Joshua~V Dillon}, {and} \bibinfo{person}{Kevin Murphy}.} \bibinfo{year}{2016}\natexlab{}.
\newblock \showarticletitle{Deep variational information bottleneck}.
\newblock \bibinfo{journal}{\emph{arXiv preprint arXiv:1612.00410}} (\bibinfo{year}{2016}).
\newblock


\bibitem[Baltrusaitis et~al\mbox{.}(2018)]%
        {baltrusaitis2018openface}
\bibfield{author}{\bibinfo{person}{Tadas Baltrusaitis}, \bibinfo{person}{A Zadeh}, \bibinfo{person}{YC Lim}, {and} \bibinfo{person}{LP Morency}.} \bibinfo{year}{2018}\natexlab{}.
\newblock \bibinfo{title}{Openface 2.0: Facial behavior analysis toolkit. En 2018 13th IEEE International Conference on Automatic Face \& Gesture Recognition (FG 2018)}.
\newblock
\newblock


\bibitem[Degottex et~al\mbox{.}(2014)]%
        {degottex2014covarep}
\bibfield{author}{\bibinfo{person}{Gilles Degottex}, \bibinfo{person}{John Kane}, \bibinfo{person}{Thomas Drugman}, \bibinfo{person}{Tuomo Raitio}, {and} \bibinfo{person}{Stefan Scherer}.} \bibinfo{year}{2014}\natexlab{}.
\newblock \showarticletitle{COVAREP—A collaborative voice analysis repository for speech technologies}. In \bibinfo{booktitle}{\emph{2014 ieee international conference on acoustics, speech and signal processing (icassp)}}. IEEE, \bibinfo{pages}{960--964}.
\newblock


\bibitem[Devlin et~al\mbox{.}(2019)]%
        {devlin2019bert}
\bibfield{author}{\bibinfo{person}{Jacob Devlin}, \bibinfo{person}{Ming-Wei Chang}, \bibinfo{person}{Kenton Lee}, {and} \bibinfo{person}{Kristina Toutanova}.} \bibinfo{year}{2019}\natexlab{}.
\newblock \showarticletitle{Bert: Pre-training of deep bidirectional transformers for language understanding}. In \bibinfo{booktitle}{\emph{Proceedings of the 2019 conference of the North American chapter of the association for computational linguistics: human language technologies, volume 1 (long and short papers)}}. \bibinfo{pages}{4171--4186}.
\newblock


\bibitem[Du et~al\mbox{.}(2023)]%
        {du2023saits}
\bibfield{author}{\bibinfo{person}{Wenjie Du}, \bibinfo{person}{David C{\^o}t{\'e}}, {and} \bibinfo{person}{Yan Liu}.} \bibinfo{year}{2023}\natexlab{}.
\newblock \showarticletitle{Saits: Self-attention-based imputation for time series}.
\newblock \bibinfo{journal}{\emph{Expert Systems with Applications}}  \bibinfo{volume}{219} (\bibinfo{year}{2023}), \bibinfo{pages}{119619}.
\newblock


\bibitem[Eyben et~al\mbox{.}(2010)]%
        {eyben2010opensmile}
\bibfield{author}{\bibinfo{person}{Florian Eyben}, \bibinfo{person}{Martin W{\"o}llmer}, {and} \bibinfo{person}{Bj{\"o}rn Schuller}.} \bibinfo{year}{2010}\natexlab{}.
\newblock \showarticletitle{Opensmile: the munich versatile and fast open-source audio feature extractor}. In \bibinfo{booktitle}{\emph{Proceedings of the 18th ACM international conference on Multimedia}}. \bibinfo{pages}{1459--1462}.
\newblock


\bibitem[Feng et~al\mbox{.}(2024)]%
        {feng2024knowledge}
\bibfield{author}{\bibinfo{person}{Xinyu Feng}, \bibinfo{person}{Yuming Lin}, \bibinfo{person}{Lihua He}, \bibinfo{person}{You Li}, \bibinfo{person}{Liang Chang}, {and} \bibinfo{person}{Ya Zhou}.} \bibinfo{year}{2024}\natexlab{}.
\newblock \showarticletitle{Knowledge-Guided Dynamic Modality Attention Fusion Framework for Multimodal Sentiment Analysis}.
\newblock \bibinfo{journal}{\emph{arXiv preprint arXiv:2410.04491}} (\bibinfo{year}{2024}).
\newblock


\bibitem[Garreau and Luxburg(2020)]%
        {garreau2020explaining}
\bibfield{author}{\bibinfo{person}{Damien Garreau} {and} \bibinfo{person}{Ulrike Luxburg}.} \bibinfo{year}{2020}\natexlab{}.
\newblock \showarticletitle{Explaining the explainer: A first theoretical analysis of LIME}. In \bibinfo{booktitle}{\emph{International conference on artificial intelligence and statistics}}. PMLR, \bibinfo{pages}{1287--1296}.
\newblock


\bibitem[Guo et~al\mbox{.}(2024)]%
        {guo2024classifier}
\bibfield{author}{\bibinfo{person}{Zirun Guo}, \bibinfo{person}{Tao Jin}, \bibinfo{person}{Jingyuan Chen}, {and} \bibinfo{person}{Zhou Zhao}.} \bibinfo{year}{2024}\natexlab{}.
\newblock \showarticletitle{Classifier-guided Gradient Modulation for Enhanced Multimodal Learning}.
\newblock \bibinfo{journal}{\emph{arXiv preprint arXiv:2411.01409}} (\bibinfo{year}{2024}).
\newblock


\bibitem[Han et~al\mbox{.}(2021)]%
        {han2021improving}
\bibfield{author}{\bibinfo{person}{Wei Han}, \bibinfo{person}{Hui Chen}, {and} \bibinfo{person}{Soujanya Poria}.} \bibinfo{year}{2021}\natexlab{}.
\newblock \showarticletitle{Improving multimodal fusion with hierarchical mutual information maximization for multimodal sentiment analysis}.
\newblock \bibinfo{journal}{\emph{arXiv preprint arXiv:2109.00412}} (\bibinfo{year}{2021}).
\newblock


\bibitem[Hazarika et~al\mbox{.}(2020)]%
        {hazarika2020misa}
\bibfield{author}{\bibinfo{person}{Devamanyu Hazarika}, \bibinfo{person}{Roger Zimmermann}, {and} \bibinfo{person}{Soujanya Poria}.} \bibinfo{year}{2020}\natexlab{}.
\newblock \showarticletitle{Misa: Modality-invariant and-specific representations for multimodal sentiment analysis}. In \bibinfo{booktitle}{\emph{Proceedings of the 28th ACM international conference on multimedia}}. \bibinfo{pages}{1122--1131}.
\newblock


\bibitem[He et~al\mbox{.}(2020)]%
        {he2020deberta}
\bibfield{author}{\bibinfo{person}{Pengcheng He}, \bibinfo{person}{Xiaodong Liu}, \bibinfo{person}{Jianfeng Gao}, {and} \bibinfo{person}{Weizhu Chen}.} \bibinfo{year}{2020}\natexlab{}.
\newblock \showarticletitle{Deberta: Decoding-enhanced bert with disentangled attention}.
\newblock \bibinfo{journal}{\emph{arXiv preprint arXiv:2006.03654}} (\bibinfo{year}{2020}).
\newblock


\bibitem[Hou et~al\mbox{.}(2025)]%
        {hou2025tf}
\bibfield{author}{\bibinfo{person}{Jingming Hou}, \bibinfo{person}{Nazlia Omar}, \bibinfo{person}{Sabrina Tiun}, \bibinfo{person}{Saidah Saad}, {and} \bibinfo{person}{Qian He}.} \bibinfo{year}{2025}\natexlab{}.
\newblock \showarticletitle{TF-BERT: Tensor-based fusion BERT for multimodal sentiment analysis}.
\newblock \bibinfo{journal}{\emph{Neural Networks}} (\bibinfo{year}{2025}), \bibinfo{pages}{107222}.
\newblock


\bibitem[Hu et~al\mbox{.}(2022)]%
        {hu2022unimse}
\bibfield{author}{\bibinfo{person}{Guimin Hu}, \bibinfo{person}{Ting-En Lin}, \bibinfo{person}{Yi Zhao}, \bibinfo{person}{Guangming Lu}, \bibinfo{person}{Yuchuan Wu}, {and} \bibinfo{person}{Yongbin Li}.} \bibinfo{year}{2022}\natexlab{}.
\newblock \showarticletitle{UniMSE: Towards unified multimodal sentiment analysis and emotion recognition}.
\newblock \bibinfo{journal}{\emph{arXiv preprint arXiv:2211.11256}} (\bibinfo{year}{2022}).
\newblock


\bibitem[Huang et~al\mbox{.}(2025)]%
        {huang2025atcaf}
\bibfield{author}{\bibinfo{person}{Changqin Huang}, \bibinfo{person}{Jili Chen}, \bibinfo{person}{Qionghao Huang}, \bibinfo{person}{Shijin Wang}, \bibinfo{person}{Yaxin Tu}, {and} \bibinfo{person}{Xiaodi Huang}.} \bibinfo{year}{2025}\natexlab{}.
\newblock \showarticletitle{AtCAF: Attention-based causality-aware fusion network for multimodal sentiment analysis}.
\newblock \bibinfo{journal}{\emph{Information Fusion}}  \bibinfo{volume}{114} (\bibinfo{year}{2025}), \bibinfo{pages}{102725}.
\newblock


\bibitem[Huang et~al\mbox{.}(2015)]%
        {huang2015bidirectional}
\bibfield{author}{\bibinfo{person}{Zhiheng Huang}, \bibinfo{person}{Wei Xu}, {and} \bibinfo{person}{Kai Yu}.} \bibinfo{year}{2015}\natexlab{}.
\newblock \showarticletitle{Bidirectional LSTM-CRF models for sequence tagging}.
\newblock \bibinfo{journal}{\emph{arXiv preprint arXiv:1508.01991}} (\bibinfo{year}{2015}).
\newblock


\bibitem[Jia et~al\mbox{.}(2022)]%
        {jia2022feature}
\bibfield{author}{\bibinfo{person}{Weikuan Jia}, \bibinfo{person}{Meili Sun}, \bibinfo{person}{Jian Lian}, {and} \bibinfo{person}{Sujuan Hou}.} \bibinfo{year}{2022}\natexlab{}.
\newblock \showarticletitle{Feature dimensionality reduction: a review}.
\newblock \bibinfo{journal}{\emph{Complex \& Intelligent Systems}} \bibinfo{volume}{8}, \bibinfo{number}{3} (\bibinfo{year}{2022}), \bibinfo{pages}{2663--2693}.
\newblock


\bibitem[Kalateh et~al\mbox{.}(2024)]%
        {kalateh2024systematic}
\bibfield{author}{\bibinfo{person}{Sepideh Kalateh}, \bibinfo{person}{Luis~A Estrada-Jimenez}, \bibinfo{person}{Sanaz~Nikghadam Hojjati}, {and} \bibinfo{person}{Jose Barata}.} \bibinfo{year}{2024}\natexlab{}.
\newblock \showarticletitle{A systematic review on multimodal emotion recognition: building blocks, current state, applications, and challenges}.
\newblock \bibinfo{journal}{\emph{IEEE Access}} (\bibinfo{year}{2024}).
\newblock


\bibitem[Khalane et~al\mbox{.}(2025)]%
        {khalane2025evaluating}
\bibfield{author}{\bibinfo{person}{Aaishwarya Khalane}, \bibinfo{person}{Rikesh Makwana}, \bibinfo{person}{Talal Shaikh}, {and} \bibinfo{person}{Abrar Ullah}.} \bibinfo{year}{2025}\natexlab{}.
\newblock \showarticletitle{Evaluating significant features in context-aware multimodal emotion recognition with XAI methods}.
\newblock \bibinfo{journal}{\emph{Expert Systems}} \bibinfo{volume}{42}, \bibinfo{number}{1} (\bibinfo{year}{2025}), \bibinfo{pages}{e13403}.
\newblock


\bibitem[Lai et~al\mbox{.}(2023)]%
        {lai2023multimodal}
\bibfield{author}{\bibinfo{person}{Songning Lai}, \bibinfo{person}{Xifeng Hu}, \bibinfo{person}{Haoxuan Xu}, \bibinfo{person}{Zhaoxia Ren}, {and} \bibinfo{person}{Zhi Liu}.} \bibinfo{year}{2023}\natexlab{}.
\newblock \showarticletitle{Multimodal sentiment analysis: A survey}.
\newblock \bibinfo{journal}{\emph{Displays}}  \bibinfo{volume}{80} (\bibinfo{year}{2023}), \bibinfo{pages}{102563}.
\newblock


\bibitem[Li et~al\mbox{.}(2021)]%
        {li2021quantum}
\bibfield{author}{\bibinfo{person}{Qiuchi Li}, \bibinfo{person}{Dimitris Gkoumas}, \bibinfo{person}{Christina Lioma}, {and} \bibinfo{person}{Massimo Melucci}.} \bibinfo{year}{2021}\natexlab{}.
\newblock \showarticletitle{Quantum-inspired multimodal fusion for video sentiment analysis}.
\newblock \bibinfo{journal}{\emph{Information Fusion}}  \bibinfo{volume}{65} (\bibinfo{year}{2021}), \bibinfo{pages}{58--71}.
\newblock


\bibitem[Liu et~al\mbox{.}(2024a)]%
        {liu2024transformer}
\bibfield{author}{\bibinfo{person}{Cong Liu}, \bibinfo{person}{Yong Wang}, {and} \bibinfo{person}{Jing Yang}.} \bibinfo{year}{2024}\natexlab{a}.
\newblock \showarticletitle{A transformer-encoder-based multimodal multi-attention fusion network for sentiment analysis}.
\newblock \bibinfo{journal}{\emph{Applied Intelligence}} \bibinfo{volume}{54}, \bibinfo{number}{17} (\bibinfo{year}{2024}), \bibinfo{pages}{8415--8441}.
\newblock


\bibitem[Liu et~al\mbox{.}(2022)]%
        {liu2022make}
\bibfield{author}{\bibinfo{person}{Yihe Liu}, \bibinfo{person}{Ziqi Yuan}, \bibinfo{person}{Huisheng Mao}, \bibinfo{person}{Zhiyun Liang}, \bibinfo{person}{Wanqiuyue Yang}, \bibinfo{person}{Yuanzhe Qiu}, \bibinfo{person}{Tie Cheng}, \bibinfo{person}{Xiaoteng Li}, \bibinfo{person}{Hua Xu}, {and} \bibinfo{person}{Kai Gao}.} \bibinfo{year}{2022}\natexlab{}.
\newblock \bibinfo{title}{Make Acoustic and Visual Cues Matter: CH-SIMS v2.0 Dataset and AV-Mixup Consistent Module}.
\newblock
\newblock
\showeprint[arxiv]{2209.02604}~[cs.MM]


\bibitem[Liu et~al\mbox{.}(2018)]%
        {liu2018efficient}
\bibfield{author}{\bibinfo{person}{Zhun Liu}, \bibinfo{person}{Ying Shen}, \bibinfo{person}{Varun~Bharadhwaj Lakshminarasimhan}, \bibinfo{person}{Paul~Pu Liang}, \bibinfo{person}{Amir Zadeh}, {and} \bibinfo{person}{Louis-Philippe Morency}.} \bibinfo{year}{2018}\natexlab{}.
\newblock \showarticletitle{Efficient low-rank multimodal fusion with modality-specific factors}.
\newblock \bibinfo{journal}{\emph{arXiv preprint arXiv:1806.00064}} (\bibinfo{year}{2018}).
\newblock


\bibitem[Liu et~al\mbox{.}(2024b)]%
        {liu2024kan}
\bibfield{author}{\bibinfo{person}{Ziming Liu}, \bibinfo{person}{Yixuan Wang}, \bibinfo{person}{Sachin Vaidya}, \bibinfo{person}{Fabian Ruehle}, \bibinfo{person}{James Halverson}, \bibinfo{person}{Marin Solja{\v{c}}i{\'c}}, \bibinfo{person}{Thomas~Y Hou}, {and} \bibinfo{person}{Max Tegmark}.} \bibinfo{year}{2024}\natexlab{b}.
\newblock \showarticletitle{Kan: Kolmogorov-arnold networks}.
\newblock \bibinfo{journal}{\emph{arXiv preprint arXiv:2404.19756}} (\bibinfo{year}{2024}).
\newblock


\bibitem[Liu et~al\mbox{.}(2024c)]%
        {liu2024sentiment}
\bibfield{author}{\bibinfo{person}{Ziyu Liu}, \bibinfo{person}{Tao Yang}, \bibinfo{person}{Wen Chen}, \bibinfo{person}{Jiangchuan Chen}, \bibinfo{person}{Qinru Li}, {and} \bibinfo{person}{Jun Zhang}.} \bibinfo{year}{2024}\natexlab{c}.
\newblock \showarticletitle{Sentiment analysis of social media comments based on multimodal attention fusion network}.
\newblock \bibinfo{journal}{\emph{Applied Soft Computing}}  \bibinfo{volume}{164} (\bibinfo{year}{2024}), \bibinfo{pages}{112011}.
\newblock


\bibitem[Liu et~al\mbox{.}(2024d)]%
        {liu2024modality}
\bibfield{author}{\bibinfo{person}{Zhizhong Liu}, \bibinfo{person}{Bin Zhou}, \bibinfo{person}{Dianhui Chu}, \bibinfo{person}{Yuhang Sun}, {and} \bibinfo{person}{Lingqiang Meng}.} \bibinfo{year}{2024}\natexlab{d}.
\newblock \showarticletitle{Modality translation-based multimodal sentiment analysis under uncertain missing modalities}.
\newblock \bibinfo{journal}{\emph{Information Fusion}}  \bibinfo{volume}{101} (\bibinfo{year}{2024}), \bibinfo{pages}{101973}.
\newblock


\bibitem[Lu et~al\mbox{.}(2024a)]%
        {lu2024hypergraph}
\bibfield{author}{\bibinfo{person}{Nannan Lu}, \bibinfo{person}{Zhiyuan Han}, {and} \bibinfo{person}{Zhen Tan}.} \bibinfo{year}{2024}\natexlab{a}.
\newblock \showarticletitle{A Hypergraph based Contextual Relationship Modeling Method for Multimodal Emotion Recognition in Conversation}.
\newblock \bibinfo{journal}{\emph{IEEE Transactions on Multimedia}} (\bibinfo{year}{2024}).
\newblock


\bibitem[Lu et~al\mbox{.}(2024b)]%
        {lu2024coordinated}
\bibfield{author}{\bibinfo{person}{Qiang Lu}, \bibinfo{person}{Xia Sun}, \bibinfo{person}{Zhizezhang Gao}, \bibinfo{person}{Yunfei Long}, \bibinfo{person}{Jun Feng}, {and} \bibinfo{person}{Hao Zhang}.} \bibinfo{year}{2024}\natexlab{b}.
\newblock \showarticletitle{Coordinated-joint translation fusion framework with sentiment-interactive graph convolutional networks for multimodal sentiment analysis}.
\newblock \bibinfo{journal}{\emph{Information Processing \& Management}} \bibinfo{volume}{61}, \bibinfo{number}{1} (\bibinfo{year}{2024}), \bibinfo{pages}{103538}.
\newblock


\bibitem[Lundberg(2017)]%
        {lundberg2017unified}
\bibfield{author}{\bibinfo{person}{Scott Lundberg}.} \bibinfo{year}{2017}\natexlab{}.
\newblock \showarticletitle{A unified approach to interpreting model predictions}.
\newblock \bibinfo{journal}{\emph{arXiv preprint arXiv:1705.07874}} (\bibinfo{year}{2017}).
\newblock


\bibitem[Madsen et~al\mbox{.}(2022)]%
        {madsen2022post}
\bibfield{author}{\bibinfo{person}{Andreas Madsen}, \bibinfo{person}{Siva Reddy}, {and} \bibinfo{person}{Sarath Chandar}.} \bibinfo{year}{2022}\natexlab{}.
\newblock \showarticletitle{Post-hoc interpretability for neural nlp: A survey}.
\newblock \bibinfo{journal}{\emph{Comput. Surveys}} \bibinfo{volume}{55}, \bibinfo{number}{8} (\bibinfo{year}{2022}), \bibinfo{pages}{1--42}.
\newblock


\bibitem[Mai et~al\mbox{.}(2019)]%
        {mai2019divide}
\bibfield{author}{\bibinfo{person}{Sijie Mai}, \bibinfo{person}{Haifeng Hu}, {and} \bibinfo{person}{Songlong Xing}.} \bibinfo{year}{2019}\natexlab{}.
\newblock \showarticletitle{Divide, conquer and combine: Hierarchical feature fusion network with local and global perspectives for multimodal affective computing}. In \bibinfo{booktitle}{\emph{Proceedings of the 57th annual meeting of the association for computational linguistics}}. \bibinfo{pages}{481--492}.
\newblock


\bibitem[Mai et~al\mbox{.}(2021)]%
        {mai2021analyzing}
\bibfield{author}{\bibinfo{person}{Sijie Mai}, \bibinfo{person}{Songlong Xing}, {and} \bibinfo{person}{Haifeng Hu}.} \bibinfo{year}{2021}\natexlab{}.
\newblock \showarticletitle{Analyzing multimodal sentiment via acoustic-and visual-LSTM with channel-aware temporal convolution network}.
\newblock \bibinfo{journal}{\emph{IEEE/ACM Transactions on Audio, Speech, and Language Processing}}  \bibinfo{volume}{29} (\bibinfo{year}{2021}), \bibinfo{pages}{1424--1437}.
\newblock


\bibitem[Mai et~al\mbox{.}(2022a)]%
        {mai2022multimodal}
\bibfield{author}{\bibinfo{person}{Sijie Mai}, \bibinfo{person}{Ying Zeng}, {and} \bibinfo{person}{Haifeng Hu}.} \bibinfo{year}{2022}\natexlab{a}.
\newblock \showarticletitle{Multimodal information bottleneck: Learning minimal sufficient unimodal and multimodal representations}.
\newblock \bibinfo{journal}{\emph{IEEE Transactions on Multimedia}}  \bibinfo{volume}{25} (\bibinfo{year}{2022}), \bibinfo{pages}{4121--4134}.
\newblock


\bibitem[Mai et~al\mbox{.}(2023)]%
        {mai2023learning}
\bibfield{author}{\bibinfo{person}{Sijie Mai}, \bibinfo{person}{Ying Zeng}, {and} \bibinfo{person}{Haifeng Hu}.} \bibinfo{year}{2023}\natexlab{}.
\newblock \showarticletitle{Learning from the global view: Supervised contrastive learning of multimodal representation}.
\newblock \bibinfo{journal}{\emph{Information Fusion}}  \bibinfo{volume}{100} (\bibinfo{year}{2023}), \bibinfo{pages}{101920}.
\newblock


\bibitem[Mai et~al\mbox{.}(2022b)]%
        {mai2022hybrid}
\bibfield{author}{\bibinfo{person}{Sijie Mai}, \bibinfo{person}{Ying Zeng}, \bibinfo{person}{Shuangjia Zheng}, {and} \bibinfo{person}{Haifeng Hu}.} \bibinfo{year}{2022}\natexlab{b}.
\newblock \showarticletitle{Hybrid contrastive learning of tri-modal representation for multimodal sentiment analysis}.
\newblock \bibinfo{journal}{\emph{IEEE Transactions on Affective Computing}} \bibinfo{volume}{14}, \bibinfo{number}{3} (\bibinfo{year}{2022}), \bibinfo{pages}{2276--2289}.
\newblock


\bibitem[Miao et~al\mbox{.}(2024)]%
        {miao2024low}
\bibfield{author}{\bibinfo{person}{Xinmeng Miao}, \bibinfo{person}{Xuguang Zhang}, {and} \bibinfo{person}{Haoran Zhang}.} \bibinfo{year}{2024}\natexlab{}.
\newblock \showarticletitle{Low-rank tensor fusion and self-supervised multi-task multimodal sentiment analysis}.
\newblock \bibinfo{journal}{\emph{Multimedia Tools and Applications}} \bibinfo{volume}{83}, \bibinfo{number}{23} (\bibinfo{year}{2024}), \bibinfo{pages}{63291--63308}.
\newblock


\bibitem[Nie et~al\mbox{.}(2024)]%
        {nie2024interpretable}
\bibfield{author}{\bibinfo{person}{Xin Nie}, \bibinfo{person}{Laurence~T Yang}, \bibinfo{person}{Zhe Li}, \bibinfo{person}{Xianjun Deng}, \bibinfo{person}{Fulan Fan}, {and} \bibinfo{person}{Zecan Yang}.} \bibinfo{year}{2024}\natexlab{}.
\newblock \showarticletitle{Interpretable Multimodal Tucker Fusion Model With Information Filtering for Multimodal Sentiment Analysis}.
\newblock \bibinfo{journal}{\emph{IEEE Transactions on Computational Social Systems}} (\bibinfo{year}{2024}).
\newblock


\bibitem[Peng et~al\mbox{.}(2022)]%
        {peng2022balanced}
\bibfield{author}{\bibinfo{person}{Xiaokang Peng}, \bibinfo{person}{Yake Wei}, \bibinfo{person}{Andong Deng}, \bibinfo{person}{Dong Wang}, {and} \bibinfo{person}{Di Hu}.} \bibinfo{year}{2022}\natexlab{}.
\newblock \showarticletitle{Balanced multimodal learning via on-the-fly gradient modulation}. In \bibinfo{booktitle}{\emph{Proceedings of the IEEE/CVF conference on computer vision and pattern recognition}}. \bibinfo{pages}{8238--8247}.
\newblock


\bibitem[Raffel et~al\mbox{.}(2020)]%
        {raffel2020exploring}
\bibfield{author}{\bibinfo{person}{Colin Raffel}, \bibinfo{person}{Noam Shazeer}, \bibinfo{person}{Adam Roberts}, \bibinfo{person}{Katherine Lee}, \bibinfo{person}{Sharan Narang}, \bibinfo{person}{Michael Matena}, \bibinfo{person}{Yanqi Zhou}, \bibinfo{person}{Wei Li}, {and} \bibinfo{person}{Peter~J Liu}.} \bibinfo{year}{2020}\natexlab{}.
\newblock \showarticletitle{Exploring the limits of transfer learning with a unified text-to-text transformer}.
\newblock \bibinfo{journal}{\emph{Journal of machine learning research}} \bibinfo{volume}{21}, \bibinfo{number}{140} (\bibinfo{year}{2020}), \bibinfo{pages}{1--67}.
\newblock


\bibitem[Rahman et~al\mbox{.}(2020)]%
        {rahman2020integrating}
\bibfield{author}{\bibinfo{person}{Wasifur Rahman}, \bibinfo{person}{Md~Kamrul Hasan}, \bibinfo{person}{Sangwu Lee}, \bibinfo{person}{Amir Zadeh}, \bibinfo{person}{Chengfeng Mao}, \bibinfo{person}{Louis-Philippe Morency}, {and} \bibinfo{person}{Ehsan Hoque}.} \bibinfo{year}{2020}\natexlab{}.
\newblock \showarticletitle{Integrating multimodal information in large pretrained transformers}. In \bibinfo{booktitle}{\emph{Proceedings of the conference. Association for Computational Linguistics. Meeting}}, Vol.~\bibinfo{volume}{2020}. NIH Public Access, \bibinfo{pages}{2359}.
\newblock


\bibitem[Retzlaff et~al\mbox{.}(2024)]%
        {retzlaff2024post}
\bibfield{author}{\bibinfo{person}{Carl~O Retzlaff}, \bibinfo{person}{Alessa Angerschmid}, \bibinfo{person}{Anna Saranti}, \bibinfo{person}{David Schneeberger}, \bibinfo{person}{Richard Roettger}, \bibinfo{person}{Heimo Mueller}, {and} \bibinfo{person}{Andreas Holzinger}.} \bibinfo{year}{2024}\natexlab{}.
\newblock \showarticletitle{Post-hoc vs ante-hoc explanations: xAI design guidelines for data scientists}.
\newblock \bibinfo{journal}{\emph{Cognitive Systems Research}}  \bibinfo{volume}{86} (\bibinfo{year}{2024}), \bibinfo{pages}{101243}.
\newblock


\bibitem[Schmidt-Hieber(2021)]%
        {schmidt2021kolmogorov}
\bibfield{author}{\bibinfo{person}{Johannes Schmidt-Hieber}.} \bibinfo{year}{2021}\natexlab{}.
\newblock \showarticletitle{The Kolmogorov--Arnold representation theorem revisited}.
\newblock \bibinfo{journal}{\emph{Neural networks}}  \bibinfo{volume}{137} (\bibinfo{year}{2021}), \bibinfo{pages}{119--126}.
\newblock


\bibitem[Shapiro(2003)]%
        {shapiro2003monte}
\bibfield{author}{\bibinfo{person}{Alexander Shapiro}.} \bibinfo{year}{2003}\natexlab{}.
\newblock \showarticletitle{Monte Carlo sampling methods}.
\newblock \bibinfo{journal}{\emph{Handbooks in operations research and management science}}  \bibinfo{volume}{10} (\bibinfo{year}{2003}), \bibinfo{pages}{353--425}.
\newblock


\bibitem[Sun et~al\mbox{.}(2024)]%
        {sun2024mfm}
\bibfield{author}{\bibinfo{person}{Shuangyang Sun}, \bibinfo{person}{Guoyan Xu}, {and} \bibinfo{person}{Sijun Lu}.} \bibinfo{year}{2024}\natexlab{}.
\newblock \showarticletitle{MFM: Multimodal Sentiment Analysis Based on Modal Focusing Model}. In \bibinfo{booktitle}{\emph{2024 IEEE International Conference on Systems, Man, and Cybernetics (SMC)}}. IEEE, \bibinfo{pages}{1524--1529}.
\newblock


\bibitem[Tishby et~al\mbox{.}(2000)]%
        {tishby2000information}
\bibfield{author}{\bibinfo{person}{Naftali Tishby}, \bibinfo{person}{Fernando~C Pereira}, {and} \bibinfo{person}{William Bialek}.} \bibinfo{year}{2000}\natexlab{}.
\newblock \showarticletitle{The information bottleneck method}.
\newblock \bibinfo{journal}{\emph{arXiv preprint physics/0004057}} (\bibinfo{year}{2000}).
\newblock


\bibitem[Tsai et~al\mbox{.}(2019)]%
        {tsai2019multimodal}
\bibfield{author}{\bibinfo{person}{Yao-Hung~Hubert Tsai}, \bibinfo{person}{Shaojie Bai}, \bibinfo{person}{Paul~Pu Liang}, \bibinfo{person}{J~Zico Kolter}, \bibinfo{person}{Louis-Philippe Morency}, {and} \bibinfo{person}{Ruslan Salakhutdinov}.} \bibinfo{year}{2019}\natexlab{}.
\newblock \showarticletitle{Multimodal transformer for unaligned multimodal language sequences}. In \bibinfo{booktitle}{\emph{Proceedings of the conference. Association for computational linguistics. Meeting}}, Vol.~\bibinfo{volume}{2019}. \bibinfo{pages}{6558}.
\newblock


\bibitem[Tsai et~al\mbox{.}(2020)]%
        {tsai2020multimodal}
\bibfield{author}{\bibinfo{person}{Yao-Hung~Hubert Tsai}, \bibinfo{person}{Martin~Q Ma}, \bibinfo{person}{Muqiao Yang}, \bibinfo{person}{Ruslan Salakhutdinov}, {and} \bibinfo{person}{Louis-Philippe Morency}.} \bibinfo{year}{2020}\natexlab{}.
\newblock \showarticletitle{Multimodal routing: Improving local and global interpretability of multimodal language analysis}. In \bibinfo{booktitle}{\emph{Proceedings of the conference on empirical methods in natural language processing. Conference on empirical methods in natural language processing}}, Vol.~\bibinfo{volume}{2020}. \bibinfo{pages}{1823}.
\newblock


\bibitem[Wang et~al\mbox{.}(2019)]%
        {wang2019deep}
\bibfield{author}{\bibinfo{person}{Qi Wang}, \bibinfo{person}{Claire Boudreau}, \bibinfo{person}{Qixing Luo}, \bibinfo{person}{Pang-Ning Tan}, {and} \bibinfo{person}{Jiayu Zhou}.} \bibinfo{year}{2019}\natexlab{}.
\newblock \showarticletitle{Deep multi-view information bottleneck}. In \bibinfo{booktitle}{\emph{Proceedings of the 2019 SIAM International Conference on Data Mining}}. SIAM, \bibinfo{pages}{37--45}.
\newblock


\bibitem[Wang et~al\mbox{.}(2025)]%
        {wang2025cime}
\bibfield{author}{\bibinfo{person}{Rui Wang}, \bibinfo{person}{Chaopeng Guo}, \bibinfo{person}{Erik Cambria}, \bibinfo{person}{Imad Rida}, \bibinfo{person}{Haochen Yuan}, \bibinfo{person}{Md~Jalil Piran}, \bibinfo{person}{Yichen Feng}, \bibinfo{person}{Xianxun Zhu}, {and} \bibinfo{person}{Mairie de Compiegne}.} \bibinfo{year}{2025}\natexlab{}.
\newblock \showarticletitle{CIME: Contextual interactionbased multimodal emotion analysis with enhanced semantic information}.
\newblock \bibinfo{journal}{\emph{The Journal of Supercomputing}} (\bibinfo{year}{2025}).
\newblock


\bibitem[Wang et~al\mbox{.}(2020)]%
        {wang2020cnn}
\bibfield{author}{\bibinfo{person}{Zijie~J Wang}, \bibinfo{person}{Robert Turko}, \bibinfo{person}{Omar Shaikh}, \bibinfo{person}{Haekyu Park}, \bibinfo{person}{Nilaksh Das}, \bibinfo{person}{Fred Hohman}, \bibinfo{person}{Minsuk Kahng}, {and} \bibinfo{person}{Duen Horng~Polo Chau}.} \bibinfo{year}{2020}\natexlab{}.
\newblock \showarticletitle{CNN explainer: learning convolutional neural networks with interactive visualization}.
\newblock \bibinfo{journal}{\emph{IEEE Transactions on Visualization and Computer Graphics}} \bibinfo{volume}{27}, \bibinfo{number}{2} (\bibinfo{year}{2020}), \bibinfo{pages}{1396--1406}.
\newblock


\bibitem[Wei and Hu(2024)]%
        {wei2024mmpareto}
\bibfield{author}{\bibinfo{person}{Yake Wei} {and} \bibinfo{person}{Di Hu}.} \bibinfo{year}{2024}\natexlab{}.
\newblock \showarticletitle{Mmpareto: boosting multimodal learning with innocent unimodal assistance}.
\newblock \bibinfo{journal}{\emph{arXiv preprint arXiv:2405.17730}} (\bibinfo{year}{2024}).
\newblock


\bibitem[Williams et~al\mbox{.}(2018a)]%
        {williams2018dnn}
\bibfield{author}{\bibinfo{person}{Jennifer Williams}, \bibinfo{person}{Ramona Comanescu}, \bibinfo{person}{Oana Radu}, {and} \bibinfo{person}{Leimin Tian}.} \bibinfo{year}{2018}\natexlab{a}.
\newblock \showarticletitle{Dnn multimodal fusion techniques for predicting video sentiment}. In \bibinfo{booktitle}{\emph{Proceedings of grand challenge and workshop on human multimodal language (Challenge-HML)}}. \bibinfo{pages}{64--72}.
\newblock


\bibitem[Williams et~al\mbox{.}(2018b)]%
        {williams2018recognizing}
\bibfield{author}{\bibinfo{person}{Jennifer Williams}, \bibinfo{person}{Steven Kleinegesse}, \bibinfo{person}{Ramona Comanescu}, {and} \bibinfo{person}{Oana Radu}.} \bibinfo{year}{2018}\natexlab{b}.
\newblock \showarticletitle{Recognizing emotions in video using multimodal DNN feature fusion}. In \bibinfo{booktitle}{\emph{Proceedings of Grand Challenge and Workshop on Human Multimodal Language (Challenge-HML)}}. \bibinfo{pages}{11--19}.
\newblock


\bibitem[Wilson and Mohan(2017)]%
        {wilson2017information}
\bibfield{author}{\bibinfo{person}{Shyju Wilson} {and} \bibinfo{person}{C~Krishna Mohan}.} \bibinfo{year}{2017}\natexlab{}.
\newblock \showarticletitle{An information bottleneck approach to optimize the dictionary of visual data}.
\newblock \bibinfo{journal}{\emph{IEEE Transactions on Multimedia}} \bibinfo{volume}{20}, \bibinfo{number}{1} (\bibinfo{year}{2017}), \bibinfo{pages}{96--106}.
\newblock


\bibitem[Wu et~al\mbox{.}(2022b)]%
        {wu2022interpretable}
\bibfield{author}{\bibinfo{person}{Jianfeng Wu}, \bibinfo{person}{Sijie Mai}, {and} \bibinfo{person}{Haifeng Hu}.} \bibinfo{year}{2022}\natexlab{b}.
\newblock \showarticletitle{Interpretable multimodal capsule fusion}.
\newblock \bibinfo{journal}{\emph{IEEE/ACM Transactions on Audio, Speech, and Language Processing}}  \bibinfo{volume}{30} (\bibinfo{year}{2022}), \bibinfo{pages}{1815--1826}.
\newblock


\bibitem[Wu et~al\mbox{.}(2022a)]%
        {wu2022characterizing}
\bibfield{author}{\bibinfo{person}{Nan Wu}, \bibinfo{person}{Stanislaw Jastrzebski}, \bibinfo{person}{Kyunghyun Cho}, {and} \bibinfo{person}{Krzysztof~J Geras}.} \bibinfo{year}{2022}\natexlab{a}.
\newblock \showarticletitle{Characterizing and overcoming the greedy nature of learning in multi-modal deep neural networks}. In \bibinfo{booktitle}{\emph{International Conference on Machine Learning}}. PMLR, \bibinfo{pages}{24043--24055}.
\newblock


\bibitem[Xiao et~al\mbox{.}(2024)]%
        {xiao2024neuro}
\bibfield{author}{\bibinfo{person}{Xiongye Xiao}, \bibinfo{person}{Gengshuo Liu}, \bibinfo{person}{Gaurav Gupta}, \bibinfo{person}{Defu Cao}, \bibinfo{person}{Shixuan Li}, \bibinfo{person}{Yaxing Li}, \bibinfo{person}{Tianqing Fang}, \bibinfo{person}{Mingxi Cheng}, {and} \bibinfo{person}{Paul Bogdan}.} \bibinfo{year}{2024}\natexlab{}.
\newblock \showarticletitle{Neuro-inspired information-theoretic hierarchical perception for multimodal learning}.
\newblock \bibinfo{journal}{\emph{arXiv preprint arXiv:2404.09403}} (\bibinfo{year}{2024}).
\newblock


\bibitem[Yang et~al\mbox{.}(2023)]%
        {yang2023confede}
\bibfield{author}{\bibinfo{person}{Jiuding Yang}, \bibinfo{person}{Yakun Yu}, \bibinfo{person}{Di Niu}, \bibinfo{person}{Weidong Guo}, {and} \bibinfo{person}{Yu Xu}.} \bibinfo{year}{2023}\natexlab{}.
\newblock \showarticletitle{Confede: Contrastive feature decomposition for multimodal sentiment analysis}. In \bibinfo{booktitle}{\emph{Proceedings of the 61st Annual Meeting of the Association for Computational Linguistics (Volume 1: Long Papers)}}. \bibinfo{pages}{7617--7630}.
\newblock


\bibitem[Yu et~al\mbox{.}(2021)]%
        {yu2021learning}
\bibfield{author}{\bibinfo{person}{Wenmeng Yu}, \bibinfo{person}{Hua Xu}, \bibinfo{person}{Ziqi Yuan}, {and} \bibinfo{person}{Jiele Wu}.} \bibinfo{year}{2021}\natexlab{}.
\newblock \showarticletitle{Learning modality-specific representations with self-supervised multi-task learning for multimodal sentiment analysis}. In \bibinfo{booktitle}{\emph{Proceedings of the AAAI conference on artificial intelligence}}, Vol.~\bibinfo{volume}{35}. \bibinfo{pages}{10790--10797}.
\newblock


\bibitem[Zadeh et~al\mbox{.}(2017)]%
        {zadeh2017tensor}
\bibfield{author}{\bibinfo{person}{Amir Zadeh}, \bibinfo{person}{Minghai Chen}, \bibinfo{person}{Soujanya Poria}, \bibinfo{person}{Erik Cambria}, {and} \bibinfo{person}{Louis-Philippe Morency}.} \bibinfo{year}{2017}\natexlab{}.
\newblock \showarticletitle{Tensor fusion network for multimodal sentiment analysis}.
\newblock \bibinfo{journal}{\emph{arXiv preprint arXiv:1707.07250}} (\bibinfo{year}{2017}).
\newblock


\bibitem[Zadeh et~al\mbox{.}(2018a)]%
        {zadeh2018memory}
\bibfield{author}{\bibinfo{person}{Amir Zadeh}, \bibinfo{person}{Paul~Pu Liang}, \bibinfo{person}{Navonil Mazumder}, \bibinfo{person}{Soujanya Poria}, \bibinfo{person}{Erik Cambria}, {and} \bibinfo{person}{Louis-Philippe Morency}.} \bibinfo{year}{2018}\natexlab{a}.
\newblock \showarticletitle{Memory fusion network for multi-view sequential learning}. In \bibinfo{booktitle}{\emph{Proceedings of the AAAI conference on artificial intelligence}}, Vol.~\bibinfo{volume}{32}.
\newblock


\bibitem[Zadeh et~al\mbox{.}(2016)]%
        {zadeh2016multimodal}
\bibfield{author}{\bibinfo{person}{Amir Zadeh}, \bibinfo{person}{Rowan Zellers}, \bibinfo{person}{Eli Pincus}, {and} \bibinfo{person}{Louis-Philippe Morency}.} \bibinfo{year}{2016}\natexlab{}.
\newblock \showarticletitle{Multimodal sentiment intensity analysis in videos: Facial gestures and verbal messages}.
\newblock \bibinfo{journal}{\emph{IEEE Intelligent Systems}} \bibinfo{volume}{31}, \bibinfo{number}{6} (\bibinfo{year}{2016}), \bibinfo{pages}{82--88}.
\newblock


\bibitem[Zadeh et~al\mbox{.}(2018b)]%
        {zadeh2018multimodal}
\bibfield{author}{\bibinfo{person}{AmirAli~Bagher Zadeh}, \bibinfo{person}{Paul~Pu Liang}, \bibinfo{person}{Soujanya Poria}, \bibinfo{person}{Erik Cambria}, {and} \bibinfo{person}{Louis-Philippe Morency}.} \bibinfo{year}{2018}\natexlab{b}.
\newblock \showarticletitle{Multimodal language analysis in the wild: Cmu-mosei dataset and interpretable dynamic fusion graph}. In \bibinfo{booktitle}{\emph{Proceedings of the 56th Annual Meeting of the Association for Computational Linguistics (Volume 1: Long Papers)}}. \bibinfo{pages}{2236--2246}.
\newblock


\bibitem[Zeng et~al\mbox{.}(2023)]%
        {zeng2023abs}
\bibfield{author}{\bibinfo{person}{Chunyan Zeng}, \bibinfo{person}{Kang Yan}, \bibinfo{person}{Zhifeng Wang}, \bibinfo{person}{Yan Yu}, \bibinfo{person}{Shiyan Xia}, {and} \bibinfo{person}{Nan Zhao}.} \bibinfo{year}{2023}\natexlab{}.
\newblock \showarticletitle{Abs-CAM: a gradient optimization interpretable approach for explanation of convolutional neural networks}.
\newblock \bibinfo{journal}{\emph{Signal, Image and Video Processing}} \bibinfo{volume}{17}, \bibinfo{number}{4} (\bibinfo{year}{2023}), \bibinfo{pages}{1069--1076}.
\newblock


\bibitem[Zeng et~al\mbox{.}(2024)]%
        {zeng2024disentanglement}
\bibfield{author}{\bibinfo{person}{Ying Zeng}, \bibinfo{person}{Wenjun Yan}, \bibinfo{person}{Sijie Mai}, {and} \bibinfo{person}{Haifeng Hu}.} \bibinfo{year}{2024}\natexlab{}.
\newblock \showarticletitle{Disentanglement translation network for multimodal sentiment analysis}.
\newblock \bibinfo{journal}{\emph{Information Fusion}}  \bibinfo{volume}{102} (\bibinfo{year}{2024}), \bibinfo{pages}{102031}.
\newblock


\bibitem[Zhang(2024)]%
        {zhang2024comprehensive}
\bibfield{author}{\bibinfo{person}{Heming Zhang}.} \bibinfo{year}{2024}\natexlab{}.
\newblock \showarticletitle{A comprehensive survey on multimodal sentiment analysis: Techniques, models, and applications}.
\newblock \bibinfo{journal}{\emph{Advances in Engineering Innovation}}  \bibinfo{volume}{12} (\bibinfo{year}{2024}), \bibinfo{pages}{47--52}.
\newblock


\bibitem[Zhang et~al\mbox{.}(2025)]%
        {zhang2025modal}
\bibfield{author}{\bibinfo{person}{Xiangmin Zhang}, \bibinfo{person}{Wei Wei}, {and} \bibinfo{person}{Shihao Zou}.} \bibinfo{year}{2025}\natexlab{}.
\newblock \showarticletitle{Modal Feature Optimization Network with Prompt for Multimodal Sentiment Analysis}. In \bibinfo{booktitle}{\emph{Proceedings of the 31st International Conference on Computational Linguistics}}. \bibinfo{pages}{4611--4621}.
\newblock


\bibitem[Zhu et~al\mbox{.}(2025)]%
        {zhu2025multimodal}
\bibfield{author}{\bibinfo{person}{Linan Zhu}, \bibinfo{person}{Hongyan Zhao}, \bibinfo{person}{Zhechao Zhu}, \bibinfo{person}{Chenwei Zhang}, {and} \bibinfo{person}{Xiangjie Kong}.} \bibinfo{year}{2025}\natexlab{}.
\newblock \showarticletitle{Multimodal sentiment analysis with unimodal label generation and modality decomposition}.
\newblock \bibinfo{journal}{\emph{Information Fusion}}  \bibinfo{volume}{116} (\bibinfo{year}{2025}), \bibinfo{pages}{102787}.
\newblock


\end{thebibliography}

\appendix
\section{Appendix}
\subsection{Datasets}
    The CMU\textminus MOSI \cite{zadeh2016multimodal} is a widely used MSA dataset. The CMU\textminus MOSI dataset consists of a total of 93 videos, each of which is divided into multiple utterances (with a maximum of 62 utterances per video). The sentiment intensity of each utterance is within the range of [\textminus 3,3], where \textminus 3 indicates the strongest negative sentiment and +3 indicates the strongest positive sentiment. To maintain consistency with previous work, we use 1,281 utterances for training, 229 utterances for validation, and 685 utterances for testing.

    The CMU\textminus MOSEI \cite{zadeh2018multimodal} is a large-scale multimodal language analysis dataset, comprising a total of 2,928 videos. Each video is segmented at the utterance level, and each utterance is annotated with two aspects: sentiment scores ranging from [\textminus 3, 3] and six basic emotions (anger, disgust, fear, joy, sadness, and surprise). We predict the sentiment scores of utterances on the CMU-MOSEI dataset. We use 16,265 utterances as the training set, 1,869 utterances as the validation set, and 4,643 utterances as the testing set.

    The CH-SIMS v2 dataset \cite{liu2022make} is a Chinese MSA dataset collected from 11 distinct scenarios including interviews, talk shows, and films to simulate real-world human-computer interactions. The videos were filtered to ensure high-quality acoustic and visual features. The dataset is partitioned into training, validation, and test sets in a 9:2:3 ratio. Specifically, the training set contains 2,722 instances, further categorized into 921 negative, 433 weakly negative, 232 neutral, 318 weakly positive, and 818 positive samples. The validation and test sets comprise 647 and 1,034 instances respectively.

\subsection{Feature Extraction Details}
    \textbf{Visual Modality:} For the CMU-MOSI and CMU-MOSEI datasets, Facet \footnote{iMotions 2017. https://imotions.com/} is used to extract a set of visual features, including facial action units, facial landmarks, head pose, etc. For the SIMSv2 dataset, OpenFace\cite{baltrusaitis2018openface} is utilized to extract facial features such as 68 facial landmarks, 17 facial action units, head pose, head orientation, and eye gaze direction. These visual features are captured from each utterance, resulting in a temporal sequence that represents the facial expressions over time.

    \textbf{Acoustic Modality:} For the CMU-MOSI and CMU-MOSEI datasets, COVAREP \cite{degottex2014covarep} is employed to extract a series of acoustic features, including pitch tracking, speech polarity, 12 Mel-frequency cepstral coefficients, glottal closure instants, and spectral envelope. For the SIMSv2 dataset, 25-dimensional eGeMAPS low-level descriptors (LLD) features are extracted using the OpenSMILE \cite{eyben2010opensmile} at a sampling rate of 16000 Hz. These features form a sequence that reflects the dynamic changes in vocal tone over the course of the utterance.

    \textbf{Language Modality:} For the CMU-MOSI and CMU-MOSEI datasets, DeBERTa\cite{he2020deberta} is used as the text embedding source for the model in this experiment, following the state-of-the-art methods \cite{xiao2024neuro}. For the SIMSv2 dataset, a pretrained BERT \cite{devlin2019bert} model, bert-base-chinese\footnote{https://huggingface.co/bert-base-chinese}, is used to learn contextual word embeddings to maintain consistency with the baseline model.

    For the CH-SIMS v2 dataset, the feature dimensions of the language, acoustic, and visual modalities are 768, 25, and 177, respectively. For CMU-MOSEI dataset, the feature dimensions of the three modalities are 768, 74, and 35, respectively. For CMU-MOSI dataset, the feature dimensions of the corresponding modalities are 768, 74, and 47, respectively.

\subsection{Training Details}
    We are developing a model on the NVIDIA RTX3090 GPU using the PyTorch framework, which includes CUDA 12.4 and torch version 2.2.2. We repeated each experiment five times to calculate the mean value of the metrics. Please refer to Table \ref{tab:hyperparameter} for detailed information on the hyperparameter settings employed in our experiments.
    
    \begin{table}[t]
    \vspace{0.1cm}
    \caption{Hyperparameter Settings of KAN-MCP.}
    \vspace{-0.1cm}
    \centering
    \resizebox{0.5\textwidth}{!}{
        \begin{tabular}{l|c|c|c}
            \hline
            \textbf{} & CMU-MOSI & CMU-MOSEI & SIMSv2 \\
            \hline
            Batch Size & 8 & 64 & 32 \\
            Text Learning Rate & 1e-5 & 1e-5 & 1e-3  \\
            Other Learning Rate & 1e-3 & 1e-2 & 1e-3 \\
            Training Epochs & 35 & 10 & 50 \\
            Time Dimensionality & 60 & 60 & - \\
            KAN Hidden Neurons & 4 & 4 & 3 \\
            Hidden Dimensionality & 3 & 3 & 3 \\
            DRD-MIB Mid Dim & 1024 & 518 & 1024 \\
            \hline
        \end{tabular}
    }
    \label{tab:hyperparameter}
    \vspace{-0.5cm}
    \end{table}

\subsection{Baselines}
    We compare the proposed KAN-MCP with the following competitive baselines:
    
    (1) \textbf{Graph Memory Fusion Network (Graph-MFN)} \cite{zadeh2018multimodal}: It effectively addresses the dynamic interaction issues in multimodal data fusion by introducing a dynamic fusion graph; 
    
    (2) \textbf{Multimodal Factorization Model (MFM)} \cite{sun2024mfm}: MFM decomposes multimodal representations into discriminative and generative factors, balancing generation and discrimination while enhancing interpretability.
    
    (3) \textbf{MultiModal InfoMax (MMIM)} \cite{han2021improving}: MMIM enhances multimodal representation by maximizing the mutual information between unimodal features and between different-level multimodal representations and unimodal features.
    
    (4) \textbf{Hybrid Contrastive Learning (HyCon)} \cite{mai2022hybrid}: HyCon explores inter - class relationships and intra-class / inter-modal interactions via intra-modal / inter-modal contrastive learning and semi-contrastive learning. 
    
    (5) \textbf{Unified MSA and ERC (UniMSE)} \cite{hu2022unimse}: UniMSE unifies the tasks of MSA and emotion recognition (ERC) in one framework. It uses a pre-trained modal fusion layer based on the T5\cite{raffel2020exploring} architecture to fuse and differentiate cross-modal representations.
    
    (6) \textbf{Contrastive Feature Decomposition (ConFEDE)} \cite{yang2023confede}: ConFEDE uses contrastive feature decomposition to break down each modality into similar and dissimilar features. It establishes contrastive relationships among all decomposed features and anchors on the similar features of text, thereby enhancing multimodal information representation.
        

    (7) \textbf{Multimodal Global Contrastive Learning (MGCL)} \cite{mai2023learning}: MGCL learns the joint representation of multimodal data from a global perspective through supervised contrastive learning, and enhances the learning effect of multimodal representation by maximizing the similarity of similar sample pairs and minimizing the similarity of dissimilar sample pairs; 

    (8) \textbf{Unimodal Label Generation and Modality Decomposition (ULMD)}\cite{huang2025atcaf}: ULMD is a multimodal sentiment analysis approach based on unimodal label generation and modality decomposition, leveraging a multi-task learning framework to improve performance by addressing redundancy, heterogeneity, and the lack of unimodal labels in existing datasets.

    (9) \textbf{Modal Feature Optimization Network (MFON)} \cite{zhang2025modal}: MFON is a novel network designed for multimodal sentiment analysis, which employs modal prompt attention mechanism, intra-modal knowledge distillation, and inter-modal contrastive learning to optimize the under-optimized visual and acoustic modal features, thereby improving the accuracy of sentiment analysis.
    
    (10) \textbf{Information-Theoretic Hierarchical Perception (ITHP)} \cite{xiao2024neuro}: ITHP uses the information bottleneck principle to construct hierarchical latent states. This compresses the information flow while retaining key pattern information, achieving effective multimodal information fusion and compression.

    (11) \textbf{Early Fusion Long Short-Term Memory (EF-LSTM)}\cite{williams2018recognizing}: EF-LSTM concatenates features from different modalities in the input phase and then uses a bidirectional long short-term memory network (Bi-LSTM \cite{huang2015bidirectional}) for sequence learning and feature-level fusion.
        
    (12) \textbf{Late Fusion Deep Neural Network (LF-DNN)}\cite{williams2018dnn}: LF-DNN separately extracts features and trains models for each modality, then fuses the prediction results of different modalities at the decision level.

    (13) \textbf{Tensor Fusion Network (TFN)} \cite{zadeh2017tensor}: TFN uses three modality embedding subnetworks and a tensor-based fusion approach. This combination aggregates unimodal,bimodal and trimodal interactions to end-to-end learn both intramodality and intermodality dynamics.
    
    (14) \textbf{Low-rank Multimodal Fusion (LMF)}\cite{liu2018efficient}: LMF produces the multimodal output representation by performing low-rank multimodal fusion with modality-specific factors.

    (15) \textbf{Memory Fusion Network (MFN)} \cite{zadeh2018memory}: MFN is a multi-view sequential learning neural model with three modules: System of LSTMs, Delta-memory Attention Network, and Multi-view Gated Memory. It integrates view - specific and cross-view information for prediction.
    
    (16) \textbf{Multimodal Transformer (MulT)} \cite{tsai2019multimodal}: MulT addresses inter-modal data misalignment and long-range dependency issues via its cross-modal attention mechanism.
    
    (17) \textbf{MISA}\cite{hazarika2020misa}: MISA decomposes modalities into modality-invariant and modality-specific features, then fuses them to predict sentiment states.
    
    (18) \textbf{Multimodal Adaptation Gate BERT (MAG-BERT)}\cite{rahman2020integrating}: MAG-BERT achieves the integration of visual and acoustic information by introducing a Multimodal Adaptation Gate.
    
    (19) \textbf{Self-Supervised Multi-task Multimodal (Self-MM)} \cite{yu2021learning}: Self-MM effectively learns modality-specific representations by auto-generating unimodal labels and jointly training on multi-modal and uni-modal tasks.
    
    (20) \textbf{Acoustic Visual Mixup Consistent (AV-MC)} \cite{liu2022make}: AV-MC module employs modality mixup to enhance non-verbal behavior representation, thereby strengthening the role of non-verbal cues in sentiment analysis.

    (21) \textbf{Knowledge-Guided Dynamic Modality Attention Fusion Framework (KuDA)}: KuDA guides multimodal feature fusion with emotional knowledge by dynamically selecting the dominant modality and adjusting the contributions of each modality.

\subsection{Information Bottleneck}
    The Information Bottleneck (IB) method aims to optimize representations by balancing complexity constraints, ensuring that latent representations are both discriminative and free of redundancy. Specifically, IB leverages mutual information (MI) to maximize the MI between the encoded representation $z$ and the target label $y$, while minimizing the MI between the encoded representation $z$ and the input $x$.

    Mutual information (MI) quantifies the amount of information obtained about one random variable through observations of another. Formally, given two random variables $x$ and $y$ with a joint distribution $p(x,y)$ and marginal distributions $p(x)$ and $p(y)$, their MI is defined as the KL divergence between the joint distribution and the product of the marginal distributions, as follows:
    
    \begin{equation}
        \begin{aligned}
        I(\mathbf{x}; y) &= I(y; \mathbf{x}) = \text{KL}\left(p(\mathbf{x}, y) \parallel p(\mathbf{x})p(y)\right) \\
        &= \int \mathrm{d}\mathbf{x} \, \mathrm{d}y p(\mathbf{x}, y) \log \frac{p(\mathbf{x}, y)}{p(\mathbf{x})p(y)} \\
        &= \mathbb{E}_{(\mathbf{x}, y) \sim p(\mathbf{x}, y)} \left[ \log \frac{p(\mathbf{x}, y)}{p(\mathbf{x})p(y)} \right]
        \end{aligned}
    \end{equation}
    
    The core objective of IB is to learn a compressed representation $z$ from the input $x$ such that $z$ is maximally discriminative for the target $y$ (i.e., maximizing $I(y;z)$). However, simply using an identity mapping (i.e., $x=z$) to achieve the highest information content is impractical, as it retains noise and redundant information that are irrelevant for prediction. To address this, IB introduces a constraint that minimizes the MI between the encoded representation $z$ and the input $x$ (i.e., $I(x;z)$). This constraint forces $z$ to retain only the discriminative information in $x$ that is relevant to predicting $y$, thereby learning a minimally sufficient representation of $x$ with respect to the label $y$. To achieve this balance, the IB optimization objective is typically defined as:
    
    \begin{equation}
        \mathcal{L}_{IB}=I(y;z)-\beta I(x;z)
    \end{equation}
    Here, $\beta\ge 0$ is a scalar that weights the trade-off between the compression constraint (minimizing redundant information) and the discriminative objective (maximizing target relevance) during optimization.

\subsection{DRD-MIB}
    Different from the traditional IB, we employ DRD-MIB to encode $x^m$ into a minimally sufficient predictive representation $z^m$, filtering out noise and redundancy while achieving feature dimensionality reduction. DRD-MIB allows individual modality representations to filter out noisy information prior to fusion and aligns the encoded unimodal distributions by applying the IB principle independently on each modality. The DRD-MIB objective function thus can be rewritten as:
    \begin{equation}
        \mathcal{L}_{DRD-MIB} = I(y;z) + \sum_{m\in\{t,a,v\}}[I(y;z^m)-\beta I(x^m; z^m)]
        \label{L-DRD-MIB}
    \end{equation}  
 
    To optimize the objective function in Eq. \ref{L-DRD-MIB}, we adopt the solution proposed in the variational information bottleneck (VIB) \cite{alemi2016deep}. For brevity, we only present the results here:
    \begin{equation}
        I(y;z)\ge\int dx dy dz p(z|x)p(y|x)p(x)\log q(y|z)
    \end{equation}
    where $q(y|z)$ serves as a variational approximation of $p(y|z)$. The entropy of the target label, i.e., $H(y)=-\int dyp(y)\log p(y)$ is irrelevant to parameter optimization and can thus be disregarded. Consequently, we optimize $I(y;z)$ by maximizing the lower bound of the objective function.

    For the second term of the DRD-MIB (i.e., the minimal information constraint):
    \begin{equation}
        I(x;z)\le \int dxdzp(x)p(z|x)\log\frac{p(z|x)}{q(z)}
    \end{equation}
    where $q(z)$ is a variational approximation to the marginal distribution $p(z)$ which is typically fixed as a standard normal Gaussian distribution. By integrating these two constraints, we derive the lower bound of the objective function:
    \begin{equation}
        \begin{aligned}
            &\mathcal{L}_{DRD-MIB} = I(y; \boldsymbol{z}) - \beta I(\boldsymbol{x}; \boldsymbol{z}) \geq \mathcal{J}_{DRD-MIB} \\
            &=\mathbb{E}_{(\boldsymbol{x}, y) \sim p(\boldsymbol{x}, y), \boldsymbol{z} \sim p(\boldsymbol{z} \mid \boldsymbol{x})} \Bigg[ \log q(y | \boldsymbol{z}) - \beta \cdot KL \left( p(\boldsymbol{z} | \boldsymbol{x}) \mid \mid q(\boldsymbol{z}) \right) \Bigg]
        \end{aligned}
        \label{drd-mib}
    \end{equation}
    where $J_{DRD-MIB}$ represents the lower bound of $L_{DRD-MIB}$. Maximizing $J_{DRD-MIB}$ enhances the lower bound of $L_{DRD-MIB}$, thereby optimizing the objective function.

    To optimize $J_{DRD-MIB}$ using deep neural networks, we posit that $p(z|x)$ follows a Gaussian distribution. Deep neural networks can be employed to learn the mean and variance of the Gaussian distribution $p(z|x)$:

    \begin{equation}
        p(\boldsymbol{z} \mid \boldsymbol{x}) = \mathcal{N} \left( \boldsymbol{\mu}(\boldsymbol{x}; \theta_\mu), \boldsymbol{\Sigma}(\boldsymbol{x}; \theta_\Sigma) \right) = \mathcal{N} \left( \boldsymbol{\mu}_z, \boldsymbol{\Sigma}_z \right)
        \label{p-zx}
    \end{equation}
    where $\mu$ parameterized by $\theta_\mu$ and $\Sigma$ parameterized by $\theta_{\Sigma}$ are the deep neural networks to learn the mean $\mu_z$ and variance $\Sigma_z$ of Gaussian distributions. However, in this way, the update of the parameters will be a problem because the computation of the gradients of parameters suffers from randomness. To resolve this problem, following \cite{wang2019deep,alemi2016deep}, we use the reparameterization trick to obtain $z$:
    \begin{equation}
        z_m = \mu_{z} + \Sigma_{z} \times \epsilon
    \end{equation}

    where $\epsilon\sim \mathcal{N}(0,I)$ denotes a standard normal Gaussian distribution, and $I$ is the identity matrix. The reparameterization trick shifts the stochasticity to $\epsilon$, enabling the explicit computation of parameter gradients. We assume that each element of the vector $z$ is independent.

    In regression tasks, the conditional distribution $q(y|z)$ is typically defined as:
    
    \begin{equation}
        q(y \mid \boldsymbol{z}) = e^{-\left\| y - \boldsymbol{D}(\boldsymbol{z}; \theta_d) \right\|_1} + C
    \end{equation}
    where $D(z,\theta_d)$ is a decoder parameterized by $\theta_d$, and $\hat y=D(z;\theta_d)$ represents the model's predicted output. $C$ is a constant term. Taking the logarithm of this distribution results in:
    \begin{equation}
        \log q(y \mid \boldsymbol{z}) = -\left\| y - \boldsymbol{D}(\boldsymbol{z}; \theta_d) \right\|_1 + C = -\left\| y - \hat{y} \right\|_1 + C
    \end{equation}

    Maximizing $q(y|z)$ in this context is equivalent to minimizing the mean absolute error (MAE) between the predicted value $\hat y$ and the true value $y$.

    Additionally, in practice, the approximate marginal distribution of the multimodal latent representation $z$ is often assumed to follow a standard normal Gaussian distribution:
    \begin{equation}
        q(z)\sim \mathcal{N}(0,I)
        \label{qz}
    \end{equation} 

    Combining Eq.\ref{p-zx} and Eq.\ref{qz}, the KL divergence term $KL\big(p(z|x)||q(z)\big)$ can be computed as: 
    \begin{equation}
        \begin{aligned}KL\left(p(\boldsymbol{z} \mid \boldsymbol{x}) \,|\, | \, q(\boldsymbol{z})\right) \ &= KL\left(\mathcal{N}(\boldsymbol{\mu}(\boldsymbol{x}; \theta_\mu), \boldsymbol{\Sigma}(\boldsymbol{x}; \theta_\Sigma)) \,|\, | \, \mathcal{N}(0, I)\right) \\&= KL\left(\mathcal{N}(\boldsymbol{\mu}_z, \boldsymbol{\Sigma}_z) \,|\, | \, \mathcal{N}(0, I)\right)\end{aligned}
    \end{equation}
    Here, we assume that the reparameterization of $p(z|x)$ and $q(z)$ allows for an analytical computation of the KL divergence. The KL divergence between two Gaussian distributions can be derived using standard calculus, and the detailed derivation is omitted here for brevity.
    
    Finally, the integral over $x$, $z$ and $y$ can be approximated by Monte Carlo sampling\cite{shapiro2003monte}. Consequently, $\mathcal{J}_{DRD-MIB}$ (Eq.\ref{drd-mib}) can be simplified as:
    \begin{equation}
        \scalebox{0.8}{$
            \begin{aligned}
                \mathcal{J}_{DRD-MIB} &\approx \frac{1}{b} \sum_{i=1}^{b} \biggl[ \log q(y_i \mid \boldsymbol{z}_i) + \\&\sum_{m\in\{t,a,v\}} \left[ \mathbb{E}_{\boldsymbol{\epsilon} \sim p(\boldsymbol{\epsilon})} \left[ \log q(y_i \mid \boldsymbol{z}_i^m) \right] - \beta \cdot KL \left( p(\boldsymbol{z}_i^m \mid \boldsymbol{x}_i^m) \, \bigg| \, \bigg| \, q(\boldsymbol{z}_i^m) \right) \right] \biggr] \\&= \frac{1}{b} \sum_{i=1}^{b} \biggl[ \log q(y_i \mid \boldsymbol{z}_i) + \sum_{m\in\{t,a,v\}} \left[ \mathbb{E}_{\boldsymbol{\epsilon} \sim p(\boldsymbol{\epsilon})} \left[ \log q(y_i \mid \boldsymbol{z}_i^m) \right] \right. \\&\left. - \beta \cdot KL \left( \mathcal{N}(\boldsymbol{\mu}_{z_i^m}, \boldsymbol{\Sigma}_{z_i^m}) \, \bigg| \, \bigg| \, \mathcal{N}(0, I) \right) \right] \biggr]
            \end{aligned}
        $}
    \end{equation}
    where $b$ denotes the sampling size (i.e., batch size), and $i$ is the index indicating each sample. By maximizing $\mathcal{J}_{DRD-MIB}$, we encourage $z$ to be maximally discriminative for the target while forgetting information about the primary multimodal representation $x$.
    
\end{document}